\documentclass{article}

\usepackage[numbers]{natbib}
\usepackage {arxiv}
\usepackage{cite}
\usepackage{floatrow}
\usepackage{amsmath,amssymb,amsfonts}
\usepackage{algorithm}
\usepackage{algorithmic}
\usepackage{textcomp}
\usepackage{xcolor}
\usepackage{float}
\usepackage{booktabs}
\usepackage{multirow}
\usepackage{enumitem}
\usepackage{setspace}
\usepackage[]{graphicx}
\usepackage{array}
\usepackage{amsmath}
\usepackage{amssymb}
\usepackage{epsfig}
\usepackage{epstopdf}
\usepackage{subfigure}
\usepackage{comment}
\usepackage{tabularx}
\usepackage{threeparttable}
\usepackage{adjustbox}
\usepackage{soul}
\usepackage{dsfont}

\begin{document}

\title {Attention versus Contrastive Learning of Tabular Data - A Data-centric Benchmarking
%
}

\author{Shourav B. Rabbani,~
        Ivan V. Medri, ~
        Manar D. Samad\\
Department of Computer Science\\
Tennessee State University\\
Nashville, TN, USA\\
\texttt{msamad@tnstate.edu} \\
}




\maketitle

\begin{abstract}

Despite groundbreaking success in image and text learning, deep learning has not achieved significant improvements against traditional machine learning (ML) when it comes to tabular data. This performance gap underscores the need for data-centric treatment and benchmarking of learning algorithms because tabular data with heterogeneous feature space differ in several ways from homogeneous image data. Recently, attention and contrastive learning breakthroughs have shifted computer vision and natural language processing paradigms. However, the effectiveness of these advanced deep models on tabular data is sparsely studied using a few data sets with very large sample sizes, reporting mixed findings after benchmarking against a limited number of baselines. We argue that the heterogeneity of tabular data sets and selective baselines in the literature can bias the benchmarking outcomes. This article extensively evaluates state-of-the-art attention and contrastive learning methods on a wide selection of 28 tabular data sets (14 easy and 14 hard-to-classify) against traditional deep and machine learning. Our data-centric benchmarking demonstrates when traditional ML is preferred over deep learning and vice versa because no best learning method exists for all tabular data sets. Combining between-sample and between-feature attentions conquers the invincible traditional ML on tabular data sets by a significant margin but fails on high dimensional data, where contrastive learning takes a robust lead. While a hybrid attention-contrastive learning strategy mostly wins on hard-to-classify data sets, traditional methods are frequently superior on easy-to-classify data sets with presumably simpler decision boundaries. To the best of our knowledge, this is the first benchmarking paper with statistical analyses of attention and contrastive learning performances on a diverse selection of tabular data sets against traditional deep and machine learning baselines to facilitate further advances in this field.

    
\end{abstract}


\section{Introduction}


Structured data in tables and relational databases, namely tabular data, are ubiquitous in applications ranging from electronic health records to banking, financing, recommendation systems, and cybersecurity~\citep{Maksims2023, Grinsztajn2022, Borisov2022, Kadra2021}. If effectively mined, these massive tabular data sources can significantly advance scientific research and positively impact the national economy, end users, and solution providers in numerous application domains. Advances in tabular data mining and informatics can be achieved by embarking on the recent deep learning revolution.  Deep learning methods have replaced traditional machine learning by innovating tailored learning architectures and algorithms for image, text, speech, audio, and video data and demonstrating unprecedented accuracy in learning hidden patterns. However, deep learning methods have not enjoyed much success over the monopoly of traditional machine learning on tabular data because boosted decision tree (BDT) methods consistently outperform deep learning methods in terms of accuracy and speed. This observation may be attributed to tabular data comprising a heterogeneous feature space with limited sample sizes. In contrast, deep learning methods are highly effective in learning homogeneous feature space with correlated sequences~\citep{Borisov2022}. 

Despite the recent advent of tabular data-specific deep-learning efforts, this area is still understudied, with sporadic observations and inconsistent findings across selective data sets and baseline methods. This inconsistency is primarily attributed to the heterogeneous structure and distribution of tabular data, unlike image or text data. Furthermore, there is a lack of established benchmarks and tabular datasets to systematically validate learning algorithms for tabular data. 

This paper identifies four deep tabular data learning strategies: 1) attention, 2) contrastive learning, 3) traditional deep learning, and 4) autoencoder with pretraining. We extensively evaluate and compare these learning strategies on a diverse set of tabular data against baseline traditional machine learning methods using several uniform evaluation metrics and statistical methods. The hypothesis is no single learning strategy can be the best choice for all tabular data sets in contrast to what is commonly observed for image and text data. In practice, machine learning methods are evaluated to identify the best method for a given domain problem. In contrast, one of the objectives of this paper is to inspire a data-centric pre-selection of learning strategy instead of using one arbitrary or trying a list of learning algorithms. 

The organization of this paper is as follows. In Section \ref{section_background} lists the preliminaries, provides justifications for deep learning strategies over traditional machine learning, and reviews recent deep learning methods proposed for tabular data. Section \ref{section_methods} elaborates on four learning strategies used for benchmarking in this paper. In Section \ref{section_expt-setup}, we present the experimental scenarios, tabular datasets, evaluation methods and metrics for benchmarking. Section \ref{section_results} narrates the interesting findings of our experiments. Section \ref{section_discussion} summarizes the key findings and provides some insights into the results with some future research directions. The paper concludes in section \ref{section_conclusions}

\begin{table*}[t]
    \caption{Contrasts between image and tabular data. The hundred most popular tabular data sets in the UCI machine learning repository~\citep{Dua_2019} are used to report the median sample size and data dimensionality.}
    \label{table_imageVStable}
    \begin{tabular}{lcc}
        \toprule
        Factors               & Image Data                          & Tabular Data                   \\ \midrule
        Homogeneity in feature space           & Yes              & Varied or mixed               \\
        Spatial Regularity    & Present                            & Absent                        \\
        Sample Size           & Large, over 50,000                 & Small, median: 660  \\
        Benchmark Data Sets   & CIFAR, MNIST                       & None                          \\
        Dimensionality        & High, over 1000                    & Low, typically around 18      \\
        Best Method           & Deep Learning & Traditional Machine Learning \\
        Special Methods       & Transfer Learning, Image Augmentation & None                        \\
        Application           & Computer Vision                    & Data Analytics                 \\ 
        \bottomrule
    \end{tabular}
\end {table*}

\section {Background}\label{section_background}
\subsection {Preliminaries} \label{section_background-preliminaries}
Tabular data can be structured in a data matrix ($X \in \Re^{n\times d}$) with $n$ samples (rows), each represented by a $d$ dimensional feature vector ${x_1, x_2, …, x_d}$ (columns). The $d-$dimensional feature space is said to be heterogeneous due to differences in scale and distribution across individual features, $p_{x_1} \neq p_{x_2} ... \neq p_{x_d}$. The presence of categorical and numerical variables in one feature space is another source of heterogeneity. In contrast, the pixel space of image data or word embeddings of text data is generally assumed to be a homogeneous feature space, potentially facilitating the deep learning process. 

Table \ref{table_imageVStable} lists the contrasts between tabular and image data to motivate the need for learning architecture and algorithms tailored to data characteristics. One obvious difference is in the sample size and data dimensionality, which are considerably low for tabular datasets. Therefore, many tabular data sets fail to effectively leverage the strength of “data-hungry” deep learning methods or make a strong case for dimensionality reduction or representation learning like image data. Therefore, deep learning methods traditionally benchmarked on image data sets may not be suitable for tabular data with heterogeneous feature space. For example, Abrar et al. show that deep learning methods improved on image benchmarks yield worse results on tabular datasets than the baseline methods\citep{ICACI_Abrar2023}. This observation infers that popular deep learning methods benchmarked on image datasets may not be suitable for non-image datasets without considering data-specific characteristics.

\subsection{Rationale for deep learning of tabular data}

There is strong evidence in recent literature that traditional machine learning outperforms deep learning on tabular data~\citep{Wang2021, Hamori2018, Kohler2019, Smith2020, Shwartz-Ziv2022, Borisov2022}. Domain researchers (e.g., health scientists) prefer traditional ML over deep learning (DL) despite the availability of large patient samples.  The monolithic choice of traditional ML disregards several cutting-edge instruments of deep learning.  First, traditional ML does not support incremental and transfer learning - a cornerstone of computational efficiency, scalability, and adaptability in modern data science. While limited sample size and data labels are major issues, their standard solutions (image augmentation, annotations by humans, and transfer learning) are non-trivial on tabular data. Second, traditional ML is not designed to learn a feature space in concordance with classification targets. Because deep learning performance is subpar on tabular data, a fusion of multimodal feature spaces (e.g., medical images, text, health record tables)~\citep{Huang2020Fusion} may not reap the full benefits of deep learning~\citep{Borisov2022}. Third, the requirement for data preprocessing steps diminishes with the availability of sizeable pre-trained image (e.g., VGG-16) or language (e.g., GPT-3) models for fine-tuning. Transfer learning via pre-trained models is not feasible in traditional ML~\citep{Gavito2023} or non-trivial for tabular data~\citep{Kim2023}. Moreover, tabular data preprocessing alone entails 50\% to 80\% of time and effort in a data science project~\citep{dataPreProcess01}, critically impacting data utility, quality, and uncertainty of model outcomes~\citep{Hancock2020}. These observations underscore the need to improve the performance of deep learning methods against the superior machine learning baselines.

\subsection{Literature review}
\label{section_literature-review}

Despite exceptional results in natural language processing and computer vision, the performance of deep learning methods is overshadowed by traditional machine learning when applied to tabular data. Tabular data have been touted as ``the last unconquered castle'' for deep learning \citep{Kadra2021}, including conclusions, such as ``deep learning is not all you need'' \citep{Shwartz-Ziv2022}. Various deep-learning solutions have been proposed for tabular data.

Attention-based methods are proposed for tabular data in two pioneering models: the Feature Tokenizer Transformer (FT-Transformer, FTT)\citep{FTT_Gorishniy2021} and Non-Parametric Transformer (NPT)\citep{NPT_Kossen2021}. FTT and NPT methods are inspired by the transformer model \citep{Vaswani2017} and convert features into memory-intensive embeddings. FTT employs attention scores to adjust embeddings, resulting in better performance on seven of eleven datasets than GBT and modified ResNet (A popular architecture for image datasets ~\citep{ResNet_he2016}). The authors of FTT have adapted ResNet for tabular data and reported a better performance than attention-based methods (AutoInt \citep{AutoInt_Song2019} and TabTransformer \citep{TabTransformer_Huang2020}) and tree-inspired methods, such as TabNet, GrowNet \citep{GrowNet_badirli2022} and NODE \citep{NODE_Popov2020}. TabTransformer creates feature embeddings only from categorical features and learns to attend between them.  Overall, FTT is a competitive baseline for future deep tabular data learning methods. Similar to FTT, NPT also learns between feature attention but goes further by learning attention between data samples. The NPT method improves classification performance on eight datasets compared to GBT, TabNet, k-nearest neighbors, and random forest. However, FTT and NPT have not been compared against each other in the literature. In other words, new deep-learning solutions for tabular data were primarily compared against traditional machine learning counterparts without comprehensively comparing against most other state-of-the-art deep-learning solutions. 

Shifting the focus to self-supervised learning, we observe noteworthy improvements when a model is pretrained by unlabeled data before supervised fine-tuning. Deep neural networks (DNN) can be configured into an autoencoder for self-supervised model pretraining~\citep{Perturbation_Abrar2022}. Abrar and Samad report that a self-supervised pretraining step yields better classification accuracy than GBT on four out of five tabular datasets~\citep{Perturbation_Abrar2022}. Recently, contrastive learning methods have shown state-of-the-art image classification performance using only one-tenth of the training samples required in standard supervised image classification. In contrastive learning, positive samples are created by corrupting image samples, then contrasted against negative image samples using the InfoNCE loss \citep{SimCLR_Chen_2020}.

Contrastive learning methods for tabular data propose new sample corruption methods for generating positive samples, such as in Self-Supervised Contrastive Learning with Random Feature Corruption (SCARF) \citep{SCARF_Bahri2021}. The SCARF method masks 60\% of the features of each data sample and replaces masked values with those obtained from a random sample, termed random feature corruption. 

Value Imputation and Mask Estimation (VIME) follow a similar random feature corruption strategy. However, the key difference in corruption is that VIME creates a mask using a Bernoulli distribution. Furthermore, VIME optimizes two loss terms:  one to reconstruct the corrupted features and the other to estimate mask vectors. The authors of SCARF argue that contrastive learning is superior, evidencing its better downstream classification performance compared to various forms of baseline autoencoder-based pretraining methods. A remarkable contribution of the SCARF method is in benchmarking multiple pretraining and contrastive learning methods on 69 tabular datasets from OpenMLCC-18 \citep{OpenMLCC18}. The SCARF-based contrastive learning method shows superior classification accuracy compared to autoencoder-based learning on 50 datasets, which is statistically significant on 24 datasets. Additionally, SCARF outperforms GBT on 41 out of 69 datasets and ranks better without statistical comparisons.

The SCARF method compares against other random feature corruption methods, especially CutMix \citep{cutmix_yun2019} and MixUp \citep{mixup_zhang2017}. CutMix replaces all masked values of each sample with values from one randomly selected sample. In contrast, SCARF randomly selects a sample for each masked value. MixUp corrupts the sample by replacing it with a linear combination of this and another randomly selected sample. Authors of the SCARF method report that other corruption methods are sensitive to feature scaling and less effective than their proposed (random feature corruption) corruption method. For example, random feature corruption is statistically better on only four of seven datasets, whereas CutMix is superior on the other three. In addition to a 4/3 win of the proposed method, the performance difference between these two methods is statistically insignificant on the remaining 62 datasets. 

Furthermore, the MixUp method is effective when the embedding space is corrupted instead of the input space \citep{Darabi2021, SAINT_Somepalli2022}.  Darabi et al. apply MixUp corruption to samples of the same pseudo label to create positive samples \citep{Darabi2021}. These pseudo labels are generated by finding k-nearest neighbors and then applying a graph-based label propagation method \citep{Iscen_2019_label_propagation}. This paper focuses on self-supervised contrastive learning methods that do not involve data labels in the learning process.

Despite extensive analysis, the SCARF method has several limitations. First, state-of-the-art attention-based methods are not compared against contrastive learning methods. Second, a comparison based on classification accuracy may not be reliable in the case of data imbalances. Third, statistical tests are limited to comparing SCARF against different variants of pretraining and contrastive learning methods. Fourth, the authors of SCARF do not share any source code to reproduce the results.

Alternatively, the embedding space can be corrupted, introducing sparsity in the network. Hajiramezanali et al. proposed this as an augmentation-free self-supervised learning strategy for tabular datasets  (STab)  \citep{STab_Hajiramezanali2022}. A sparsity mask is generated randomly via a Bernoulli distribution. The authors minimize the cosine distance between pairs of outputs of the same input instead of using infoNCE loss. Their model is superior to SCARF on four tabular datasets: Income, Gesture, Robot, and Theorem. However, STab is not a contrastive learning method without the corruption of feature space and, hence, is excluded from our experiment.

A few methods have proposed self-attention and between-sample attention within the contrastive framework. For example, Somepalli et al. propose Self-Attention and Intersample Attention Transformer (SAINT) \citep{SAINT_Somepalli2022}. SAINT masks each feature value with a probability $p$ following a binomial distribution (p=0.3) and uses CutMix to replace the masked values. In addition to corrupting the input feature space, they also corrupted the feature embeddings using MixUp in their proposed contrastive learning framework.  Additionally, they use self-attention (between features) and inter-sample (between samples) attention in their encoder subnetwork. The authors report superior performance on ten of 14 data sets against traditional ML baselines, VIME, TabNet, and TabTransfomer. 

In addition, Ucar et al. \citep{SubTab_Ucar2021} propose Subsetting Features of Tabular Data for Self-Supervised Representation Learning (SubTab). SubTab creates multiple versions of the dataset by taking subsets of features and adding noise to generate positive and negative sample pairs for contrastive learning. The authors report superior performance on all five tabular data sets compared to Random Forest, XGBoost, Autoencoders, and VIME.  The authors of STab report that SCARF performs similarly to SubTab \citep{STab_Hajiramezanali2022}. Therefore, we exclude SubTab from our list of baseline methods. 

Very recently, ExcelFormer has been benchmarked against traditional machine learning methods (XGBoost and CatBoost) on 12 selected classification datasets~\citep{Chen2023_ExcelFormer}. All 12 datasets yield a near-perfect classification performance score (e.g., 0.9993-0.9996), indicating the presence of easy-to-classify decision boundaries. These datasets allow a minimal scope to do 
meaningful statistical comparisons between methods.  

TabPFN is another model tested on 30 selectively small datasets from OpenML-CC18 against Gradient-boosting tree-based methods~\citep{hollmann2022_tabpfn}. Here, three datasets yield near-perfect classification performance, and 23 datasets show an improvement of less than 0.01 in the area under the receiver operating characteristic curve (AUC-ROC). TapCaps is evaluated on only eight datasets\citep{chen2023_tabcaps}. The difference in classification performance between the proposed and baseline methods is 0.01 or less.

The T2G-Former method is evaluated on eight classification datasets \citep{yan2023_t2g}. The difference in classification accuracy is within 0.005 on two datasets and below 0.01 on four datasets. Moreover, the proposed method fails to outperform XGBoost on three datasets and shows a 6\% improvement in classifying one dataset compared to other baselines. Overall, these articles report very close classification accuracies across the methods, and without statistical tests, it is hard to ascertain their competitive advantage over baseline methods.

A recent survey article on deep learning methods proposed for tabular data uses only five datasets to benchmark 12 deep learning methods, including tree- and attention-based methods such as SAINT, while excluding more recent methods (e.g., FTT, NPT, and SCARF)~\citep{Borisov2022}. Grinsztajn et al. have compared the model performance by tuning the hyperparameters of only four neural network-based methods (FTT, SAINT, ResNet, MLP) against four traditional machine learning methods (Random forest, XGBoost, Gradient Boosting Tree, HistGradientBoosting Tree)
using 22 tabular datasets. One important survey article on deep learning of tabular data uses five selective datasets with very large sample sizes (10k, 20k, 32k, 581k, and 11M)~\citep{Borisov2022}. The FTT method is evaluated on datasets with more than 20k samples~\citep{FTT_Gorishniy2021}. Newer methods such as T2G-Former \citep{yan2023_t2g} and Excel-Former \citep{Chen2023_ExcelFormer} are evaluated on large datasets with a minimum sample size of 10k and 6k, respectively. Such selective tabular datasets with large sample sizes are ideal for deep learning but rare in practice. As shown in Table~\ref{table_imageVStable}, most tabular datasets include several hundreds of samples, not in the order of tens of thousands. Therefore, the effectiveness of proposed deep learning models on tabular datasets of practical size and dimensionality is not well known. In contrast to recent survey articles on deep learning of tabular data, presenting experiments on selective methods and datasets \citep{Borisov2022, Grinsztajn2022}, we take a new data-centric approach to benchmarking involving statistical comparisons on a large spectrum of tabular data. This article goes beyond investigating the effectiveness of 
deep learning on tabular data against traditional machine learning to focus on the performance of attention and contrastive learning in this data domain. 

\section {Methods} \label{section_methods}
This section presents thirteen methods we test on tabular datasets. These methods are grouped into four categories:  (1) baseline neural networks: fully connected Deep Neural Networks (DNN) and DNN with Autoencoder pretraining (DNN-AE), (2) attention-based neural networks: TabNet \citep{TabNet_Arik2021}, FT-Transformer (FTT) \citep{FTT_Gorishniy2021}, Non-Parametric Transformer (NPT) \citep{NPT_Kossen2021}, and Self-Attention and Intersample Attention Transformer (SAINT) \citep{SAINT_Somepalli2022}, (3) contrastive learning methods (SCARF~\citep {SCARF_Bahri2021}) using five corruption strategies (pass, additive noise, sampling from feature distribution, random feature corruption (RF), and CutMix proposed in~\citep{cutmix_yun2019}, and (4) traditional machine learning: Logistic Regression (LR) and Gradient Boosting Decision Trees (GBT).

\begin{table*}[]
\caption{Summary of tabular datasets used in this study. FS-ratio = Feature-sample ratio. Cat-features = categorical features. A dataset is hard if gradient boosting classifier outperforms logistic regression by 4\% or more, otherwise, it is easy. The c-score is the mean of absolute correlations of all feature pairs.}
\vspace{-0pt}
\scalebox{0.8}{
\begin{tabular}{llccccccc}
\toprule
{OpenML Id} & Name &  Samples &  Features &  Cat-features &  Classes &  FS-ratio (\%)  & Difficulty  & C-score \\ \midrule
4538 & Gesture-phase-segmentation-processed & 9873 & 32 & 0 & 5 & 0.32 & Hard & 0.09 \\
40975 & Car & 1728 & 6 & 6 & 4 & 0.38 & Hard & 0.05 \\
40701 & Churn & 5000 & 20 & 4 & 2 & 0.40 & Hard & 0.03 \\
1497 & Wall-robot-navigation & 5456 & 24 & 0 & 4 & 0.44 & Hard & 0.17 \\
469 & Analcatdata\_dmft & 797 & 4 & 4 & 6 & 0.50 & Easy & 0.09 \\
1464 & Blood-transfusion-service-center & 748 & 4 & 0 & 2 & 0.54 & Hard & 0.47 \\
23 & Cmc & 1473 & 9 & 7 & 3 & 0.61 & Hard & 0.13 \\
11 & Balance-scale & 625 & 4 & 0 & 3 & 0.64 & Easy & 0.00 \\
1475 & First-order-theorem-proving & 6118 & 51 & 0 & 6 & 0.83 & Hard & 0.24 \\
50 & Tic-tac-toe & 958 & 9 & 9 & 2 & 0.94 & Easy & 0.09 \\
1067 & Kc1 & 2109 & 21 & 0 & 2 & 1.00 & Hard & 0.72 \\
37 & Diabetes & 768 & 8 & 0 & 2 & 1.04 & Easy & 0.17 \\
40982 & Steel-plates-fault & 1941 & 27 & 0 & 7 & 1.39 & Hard & 0.25 \\
1480 & Ilpd & 583 & 10 & 1 & 2 & 1.72 & Easy & 0.18 \\
1068 & Pc1 & 1109 & 21 & 0 & 2 & 1.89 & Hard & 0.63 \\
46 & Splice & 3190 & 61 & 60 & 3 & 1.91 & Easy & 0.04 \\
31 & Credit-g & 1000 & 20 & 13 & 2 & 2.00 & Easy & 0.06 \\
54 & Vehicle & 846 & 18 & 0 & 4 & 2.13 & Easy & 0.41 \\
1050 & Pc3 & 1563 & 37 & 0 & 2 & 2.37 & Hard & 0.41 \\
1049 & Pc4 & 1458 & 37 & 0 & 2 & 2.54 & Hard & 0.35 \\
1487 & Ozone-level-8hr & 2534 & 72 & 0 & 2 & 2.84 & Hard & 0.35 \\
40994 & Climate-model-simulation-crashes & 540 & 20 & 0 & 2 & 3.70 & Easy & 0.01 \\
1494 & Qsar-biodeg & 1055 & 41 & 0 & 2 & 3.89 & Easy & 0.17 \\
1063 & Kc2 & 522 & 21 & 0 & 2 & 4.02 & Easy & 0.78 \\
1510 & Wdbc & 569 & 30 & 0 & 2 & 5.27 & Easy & 0.39 \\
458 & Analcatdata\_authorship & 841 & 70 & 0 & 4 & 8.32 & Easy & 0.15 \\
1485 & Madelon & 2600 & 500 & 0 & 2 & 19.23 & Hard & 0.02 \\
4134 & Bioresponse & 3751 & 1776 & 0 & 2 & 47.35 & Easy & 0.09 \\
\bottomrule
\end{tabular}
}
\vspace{-10pt}
\label{table_dataset}
\end{table*}

\subsection {Tabular datasets}
 \label{section_methods-tabular}
Tabular datasets can vary widely in statistics, variable types, and decision boundaries required for discriminant analysis. Unlike image data sets, the heterogeneity of tabular data and feature space can widely vary the classification performance of a learning algorithm. Therefore, we argue that the model performance is primarily dataset-dependent, and the selection of datasets can bias benchmarking. In other words, there is no single best model for all tabular datasets.  For example, tabular datasets showing almost perfect classification accuracy (~99.9\%) using a logistic regression may overfit deep learning models, resulting in inferior accuracies~\citep{Grinsztajn2022}. Therefore, similar easy-to-classify datasets are not suitable candidates for deep learning.

In contrast, we define hard-to-classify datasets as those showing a 4\% or higher improvement in accuracy when a non-linear GBT classifier is used instead of logistic regression. We have adapted this idea from Grinsztajn et al. \citep{Grinsztajn2022}. We identify only 14 tabular datasets that satisfy the definition of the hard-to-classify category. Likewise, 14 easy-to-classify tabular data sets are selected for a fair comparison, totaling 28 tabular datasets. Table \ref{table_dataset} summarizes the size, dimensions, number of categorical variables, and number of classification targets of the selected datasets. Additionally, we show two metrics to define data statistics: (i) feature-sample ratio (FS-ratio) and (ii) mean of 
pairwise absolute correlations (C-score) computed as $mean~\{|\rho_{ij}| : i<j\}$, where $\rho_{ij}$ is the correlation between features $i$ and $j$.

\subsection{Baseline neural networks}
\subsubsection{DNN} 
Deep neural networks (DNN) are fully connected networks mapping input data to multiple layers of non-linear transformations to achieve classification at the final layer. Each layer consists of a ReLU activation layer for non-linear transformation and is trained using a cross-entropy loss. A fully-connected DNN is the default choice for deep learning of tabular data. 


\subsubsection{Pretraining via an autoencoder} 
A self-supervised pretraining step has substantially improved the downstream classification accuracy over training a DNN from scratch~\citep{Perturbation_Abrar2022}. In a self-supervised pertaining setting, an autoencoder maps input data to a low-dimensional latent space, which is then trained to decode or reconstruct the input from the latent space. A data reconstruction loss estimating the mean squared error between the input and reconstructed data is minimized by training an autoencoder. This self-supervised learning provides an effective initialization of the model's trainable parameters. The trained encoder part is coupled with a classifier layer after replacing the decoder part for downstream classification (fine-tuning and testing).

\begin{figure*}[t]
\centering
\vspace{-10pt}
 \includegraphics[trim=0cm 0cm 0cm 0cm, width=1\textwidth] {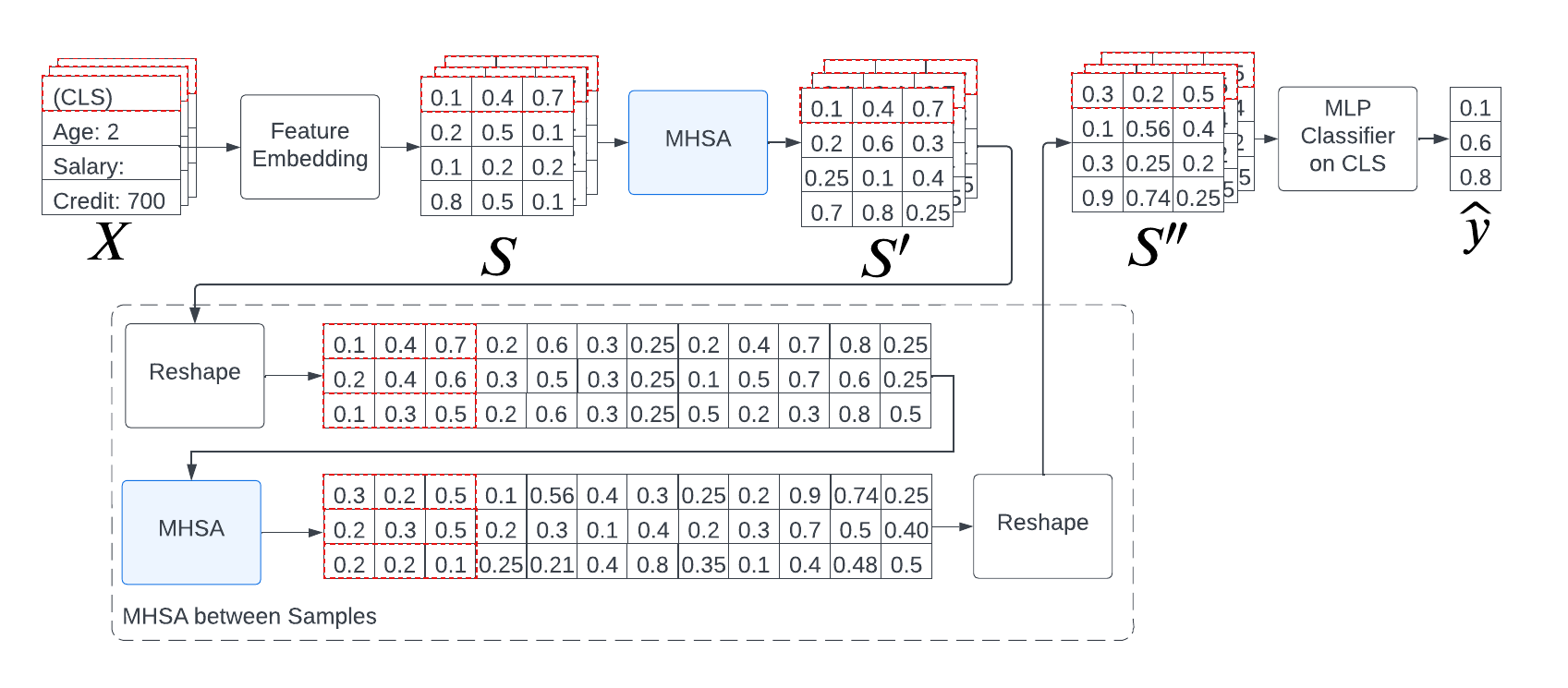}
\vspace{-10pt}
\caption {Attention between-feature and between-sample with a classifier head. Here, MHSA is the multi-headed self-attention \citep{Vaswani2017}. $S$ is the feature embedding including the CLS token~\citep{FTT_Gorishniy2021, NPT_Kossen2021, SAINT_Somepalli2022}. Between-feature attention improves $S$ to $S'$, and between-sample attention further improves it to $S''$. The final embeddings of the CLS tokens are streamlined in a classifier head to generate class logits.} 
\label{figure_attention}
\end{figure*}

\subsection{Attention-based learning}
 
Attention is not a new concept~\citep{Bahdanau2014, cho2014rnn, luong2015globalAttention, Chaudhari2021attentionSurvey1, Niu2021attentionSurvey2}, its prominence rose with the introduction of the Transformer architecture~\citep{Vaswani2017}. Transformers, initially designed for natural language translation tasks, have later found their application in other data applications, including images \citep{Dosovitskiy2020Vit} and tabular data \citep{TabTransformer_Huang2020, AutoInt_Song2019, FTT_Gorishniy2021, NPT_Kossen2021, SAINT_Somepalli2022}. A crucial step in adapting the Transformer to tabular data involves transforming column features to mimic word embeddings, where each feature is assumed as a word. This process, termed Feature-Tokenization by Gorishniy et al. \citep{FTT_Gorishniy2021}, linearly transforms each feature using trainable weights by updating the input shape from $(n \times d)$ to $(n \times d \times k)$, where $k$ is the embedding dimension for each of the $d$ features in a data table.

The introduction of an additional feature, the classification token (CLS), inspired by its use in natural language processing \citep{devlin2018bert}, is pivotal. CLS is input to an MLP subnetwork for classification or regression tasks, increasing the final shape of $S$ to $(n \times d+1 \times k)$. Subsequently, Multi-Head Self-Attention (MHSA) is applied to feature embeddings, facilitating attention between $d+1$ features within a sample. Next, the feature embeddings are reshaped to $(1 \times n \times (d+1)*k)$ to compute between-sample attention using another MHSA module. The result is then reshaped back to $(n \times d+1 \times k)$. This mechanism updates the CLS through attention, as depicted as a simplified attention-based classifier shown in Figure \ref{figure_attention}.

\subsubsection{TabNet~\citep{TabNet_Arik2021}} TabNet is one of the earliest deep learning solutions for tabular data, which selects important features similar to ensemble decision trees via sequential decision steps. Each step uses an attention mechanism to select a subset of features for learning an embedding and passes the feature subset to the next step. In subsequent steps, the feature selection process repeats on selected features from the previous step to generate new embeddings. All embeddings are aggregated using a linear layer for downstream classification. A sparse entropy loss is obtained following feature selection in each decision step. The model is trained until the total loss, consisting of a prediction loss and the sparse entropy loss, converges.

\subsubsection{FT-Transformer~\citep{FTT_Gorishniy2021}} Feature Tokenizer + Transformer (FT-Transformer) brings multiheaded self-attention (MHSA) (attention between features) to learn tabular data. The embedding generated by the attention mechanism is used to classify target labels. The supervised model is trained using a cross-entropy loss. 

\subsubsection{NPT~\citep{NPT_Kossen2021}}  Non-parametric Transformer (NPT) uses attention between variables as well as between samples to predict target labels. The model is trained to predict stochastically masked entries in tabular data to learn the relationship between variables and samples. The model is trained to minimize the reconstruction loss of masked features and cross-entropy loss on the target labels. 

\subsubsection{SAINT~\citep{SAINT_Somepalli2022} }Self-Attention and Intersample Attention Transformer (SAINT) shares an architecture similar to NPT that learns attention between samples and between features.  Additionally, the SAINT method involves a pre-training stage based on contrastive learning~\citep{SimCLR_Chen_2020} and standard data reconstruction loss. Details on contrastive learning methods are provided in the next section.

\subsection{Contrastive learning}\label{section_contrastive}
Self-supervised contrastive learning methods are proposed for learning tabular data inspired by their remarkable success in image classification~\citep{SimCLR_Chen_2020, grill2020bootstrap, chen2020improved, tian2020goodviews, chen2020big, zbontar2021barlow, he2020momentum, wang2020understanding}. Image augmentation via image rotation, scaling, and cropping generates positive samples for learning transformation-invariant feature space in image classification. An image sample is corrupted by two transformations to create a pair of positive samples. Negative pairs are created by pairing two different image samples or their corrupted versions. The positive and negative image pairs are used to train a neural network by optimizing an InfoNCE contrastive loss~\citep{SimCLR_Chen_2020} 

\begin{equation}\label{equation_simclr_1}
    \mathcal{L}  = \frac{1}{2N}\sum_{(i,j)\in P} -log 
    \frac{exp(s_{i,j} / \tau)}{\sum^{2N}_{k=1} \mathds{1}_{[k \neq i]} \ exp(s_{i,k} / \tau)},
\end{equation}
where $P$ is the set of all positive pairs, $s_{i,j}$ is the cosine similarity between the neural network-generated embeddings of a pair of positive samples, and $\mathds{1}_{[k \neq i]}$ is an indicator function that returns $1$ when $k$ is not equal to $i$ otherwise 0. The numerator in Equation \ref{equation_simclr_1} includes similarities between positive pairs, whereas the denominator includes all possible pairs. The parameter $\tau$ is the temperature to scale the similarity and is set as 1. Each term of the loss in Equation 1 can be interpreted as the negative log of the softmax function when applied to all pairs of samples $(i,j)$ with $j\neq i$. The pair $(i, i)$ is not considered in contrastive learning for being identical. Therefore, contrastive learning aims to obtain a feature space or embedding that pulls similar or positive samples closer and repels the negative samples farther.

\begin{figure}[t]
\centering
 \includegraphics[trim=0cm 0cm 0cm 0cm, width=0.7\textwidth] {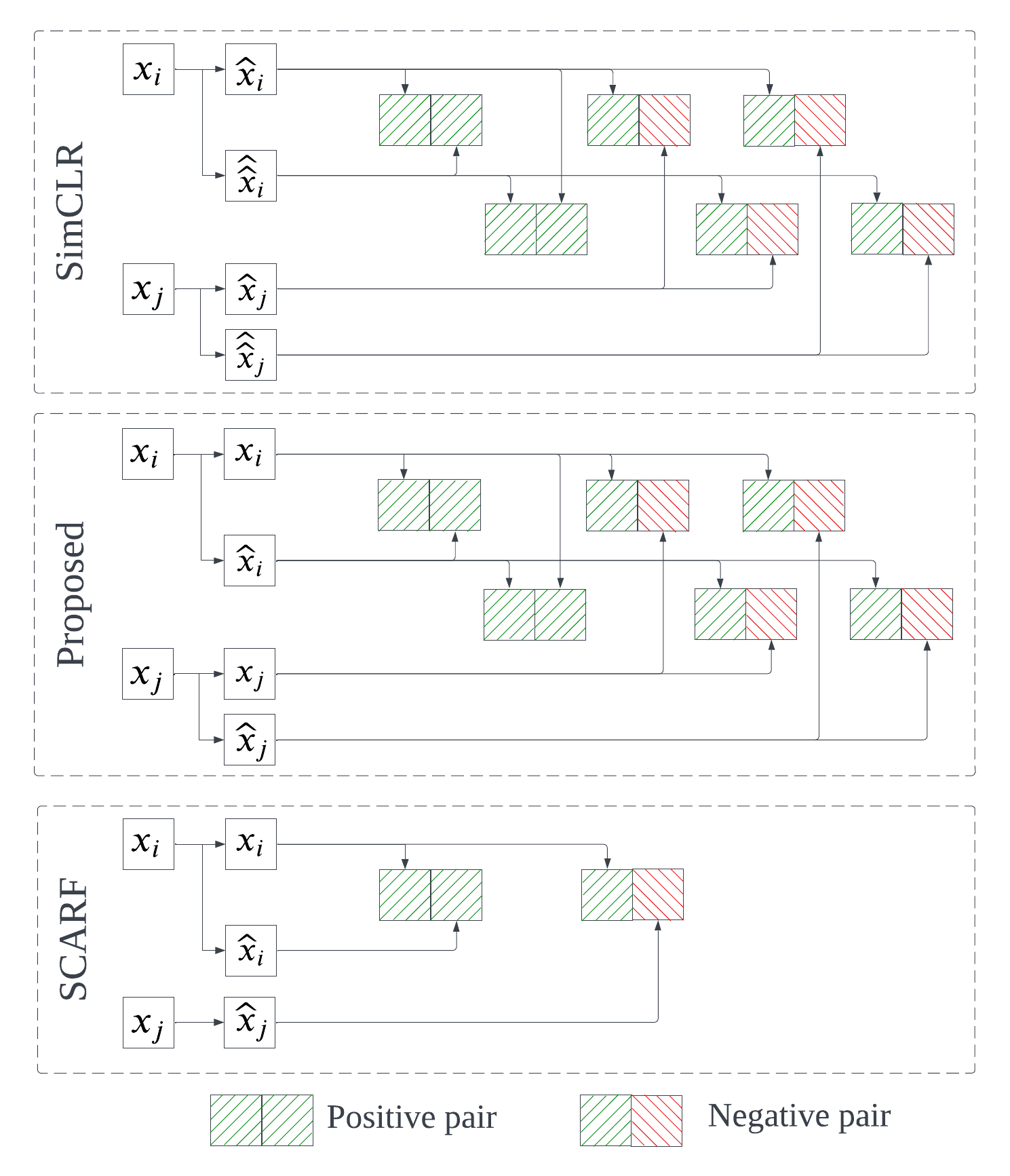}
\vspace{-10pt}
\caption {Formation of positive and negative pairs in SimCLR, SCARF, and our proposed method. Here, $x_i$ and  $x_j$ denote two different samples where ($i \neq j$) and $\widehat{x}_i$, $\widehat{x}_j$ are the corrupted versions of original samples.} 
\label{figure_corruption-similarity}
\end{figure}

However, standard image corruption methods are not trivial on tabular data. Recent work on contrastive learning of tabular data focuses on contributing new corruption methods to prepare positive samples~\citep{SCARF_Bahri2021, VIME_Yoon2020, SAINT_Somepalli2022, Rubachev2022, MIDA_Gondara2018, TABBIE_Iida2021}. In tabular data, data augmentation is achieved by corrupting samples in two steps. First, a binary mask is created to mark the candidate values for corruption. Second, a filling strategy is adopted to replace or corrupt these candidate values. Two copies (corrupted ($\hat {x}_i$) and uncorrupted ($x_i$)) of the same sample are obtained as a pair of positive samples. Additionally, we identify two advantages of corrupting one sample instead of both in a positive pair. First, the feature representation of the input (uncorrupted sample) is available via self-supervised pretraining. Second, the computation time of the loss is substantially reduced due to fewer corrupting (masking and filling) requirements. On the other hand, the negative sample pairs constitute two different samples ($x_i$,~$x_j$, $i \neq j$), such as a pair of their corrupted or uncorrupted versions ($\hat {x}_i$,~$x_j$), ($\hat {x}_i$,~$\hat {x}_j$), ($x_i$,~$\hat {x}_j$), ($x_i$,~$x_j$). Figure~\ref{figure_corruption-similarity} compares the pairing of corrupted and uncorrupted samples for image contrastive learning (standard SimCLR), a state-of-the-art contrastive learning method for tabular data (SCARF), and our proposed contrastive learning. In addition to the negative pairs used in SCARF  ($x_i$,~$\hat {x}_j$), our proposed contrastive learning includes uncorrupted pairs $(x_i, ~ x_j, i\neq j)$ and corrupted versions ($\hat {x}_i$,~$\hat {x}_j$, i$\neq$ j) as negative sample pairs. 

SCARF is a state-of-the-art contrastive learning method for tabular data~\citep{SCARF_Bahri2021}. In SCARF, given a dataset $M$, positive samples are created using only one corrupted version $\hat{M}$ of the dataset, which is achieved by masking 60\% of the features in every sample (row) and filling those masked values using different strategies (See Table \ref{table_corruption}). Table \ref{table_corruption} presents four masking and nine filling strategies used to corrupt tabular data. MIDA uses a data corruption method to facilitate the training of a denoising autoencoder~\citep{MIDA_Gondara2018}. VIME trains a similar neural network to identify and recover corrupted input data created using RFC-based corruption~\citep{VIME_Yoon2020}.

 \begin{table*}[t]
 \caption{Feature corruption strategies proposed for generating positive sample pairs for contrastive learning. Similar filling strategies are enumerated using the same number. The methods with $\star$ are proposed in the corresponding paper.}
 \vspace{-5pt}
\scalebox{0.73}{
\begin{tabular}{lll} \toprule
Article & Masking strategy & Filling strategy \\ \midrule
SCARF \citep{SCARF_Bahri2021} & 60\% of the columns in every sample 
   & Filling each masked value with zero (i) \\
 & & Filling each masked value with mean \\
 & & $\star$ Filling each masked value from a random sample (ii)\\ 
 & & Filling every masked value from a random sample (CutMix) \citep{cutmix_yun2019} (iii)\\
 & & Mixing every masked value with a random sample (MixUp) \citep{mixup_zhang2017} \\ 
 & & Adding Gaussian noise $N(0, 0.5^2)$ to each masked value (iv) \\ 
 & & No corruption \\ \midrule
SubTab \citep{SubTab_Ucar2021} & Random block of neighboring feature columns; 
   & $\star$Using only masked values \\
 & Masking each value using Binomial distribution  
   & Filling each masked value with zero (i) \\ 
 & & Filling each masked value with a random sample (ii) \\ 
 & & Adding Gaussian noise $N(0, 0.5^2)$ to each masked value (iv) \\ 
 \midrule
VIME \citep{VIME_Yoon2020} & \begin{tabular}[t]{@{}l@{}}Masking using Bernoulli distribution of probability p. \\ p selected from a validation set\end{tabular} 
   & Filling every masked value from a random sample (CutMix) \citep{cutmix_yun2019} (iii) \\ \midrule
MIDA \citep{MIDA_Gondara2018} & Masking 20\% data with MCAR/MNAR/MAR 
   & Filling every masked value with zero (i) \\ \midrule
SAINT \citep{SAINT_Somepalli2022} & Masking with Bernoulli distribution (p=0.3) 
   & Filling every masked value with a random sample (CutMix) \citep{cutmix_yun2019} (iii) \\ \midrule
Rubachev et al. \citep{Rubachev2022} & 60\% of all columns in every sample 
   & Filling each masked value from a random sample of different target labels \\ \bottomrule
\end{tabular}
\label{table_corruption}
}
\vspace{-10pt}
\end{table*}

For contrastive methods, we investigate the effectiveness of five corruption strategies shown in Table~\ref{table_corruption} as follows.

\begin{itemize}
    \item Pass: No values are altered.
    \item Noise: A value sampled from Gaussian distribution $\mathcal{N}(0, 1^2)$ is added to each masked value. Corruption is done after preprocessing the features.
    \item Sample: Features are assumed to be normally distributed with feature mean ($\mu$) and standard deviation ($\sigma$). Each masked value is replaced by a value sampled from corresponding feature distribution $\mathcal{N}(\mu, \sigma^2)$. Corruption is done after preprocessing the features.
    \item CutMix: Masked values of a sample are replaced using corresponding feature values of another random sample. This is the most commonly used data corruption method in contrastive learning of tabular data~\citep{cutmix_yun2019, SAINT_Somepalli2022, SCARF_Bahri2021}. 
    \item Random Feature Corruption (RFC) \citep{SCARF_Bahri2021}: Each masked value in a sample is replaced with the same feature value taken from another random sample. The difference between RFC and CutMix is that CutMix selects a random sample to replace all the masked values of a given sample. In contrast, RFC replaces masked values of a sample using values taken from different random samples.
    \item Proposed Within Cluster Replace (WCR): Given a data set with $K$ classes, k-mean clustering is used to obtain $K$ clusters. Instead of using a mask, all feature values of a sample are replaced using the corresponding feature values of another random sample within the same cluster. 
\end{itemize}

\section{Experimental setup and model evaluation}

The sections below present the experimental steps and evaluation metrics used in this paper. 

\subsection{Experimental setup} \label{section_expt-setup}

We randomly sample data 30 times to create train-validation-test splits to ensure enough samples for statistical tests. Every time, 70\% samples are used for training, 10\% are used for model validation, and the remaining 20\% are set apart for testing. A predefined random state seed for data split reproduces the same train, validation, and test samples for all experiments.

Our baseline DNN and DNN-AE models have the same number of layers as the SCARF network. DNN has four layers with a decreasing number of neurons and two layers for classification head, where each hidden layer uses ReLU activation. The fully connected neural network architecture appears as input-256-128-64-32-32-output. Accordingly, the autoencoder has an input-256-128-64-32-64-128-256-input architecture during pretraining.

We use the default hyperparameter setting for all deep tabular models with minimal changes to the source code. However, hyperparameters are tuned for the traditional machine learning methods, as done in practice. All neural network-based models are trained using mini-batch gradient descent with a batch size of 128 and an ADAM optimizer with a learning rate of 0.001. We train the models for 1000 epochs and perform an additional 200 epochs of finetuning for models that require pretraining. We search and save the best model with the lowest validation loss during self-supervised pre-training for the subsequent fine-tuning step.

Contrastive models are pre-trained using the infoNCE loss \citep{SimCLR_Chen_2020} with temperature 1. Positive and negative pairs are created according to the methods explained in Section \ref{section_contrastive} and depicted in Figure \ref{figure_corruption-similarity}.

We follow SCARF's validation method involving a cyclic approach. For each corruption strategy, we create ten replicas of the validation set. These replicas are then corrupted to form ten distinct validation sets, which remain unchanged throughout the training process. After each epoch, a different corrupted validation set is chosen cyclically to validate the model. For fine-tuning, we replace the projector head with a classification head. The network is then trained in a supervised manner using the cross-entropy loss.

\subsection{Model evaluation}\label{section_model-evaluation}

The SCARF method uses Welch's t-test to statistically compare the classification accuracies of different methods, which is summarized in a win matrix. It is well known that model accuracies can be biased due to class imbalance. Therefore, we use F1-scores to compare the classification performances. Instead of the Welch t-test, we use the Wilcoxon signed rank test because this it does not require any normality assumption as t-tests and is preferred in statistically comparing score data (e.g., F1 scores)\citep{Maksims2023, Scheff2016, Samad2019JAAC}.

\textbf{Weighted F1 score.} The classification performance of each method on each dataset is reported using an average weighted F1 score and standard deviation obtained across 30 bootstrapped test sets. Methods that do not produce results due to ``Out of Memory'' exceptions are reported as ``OOM''.

\textbf{Average rank.} It has been shown in previous studies that one method does not perform the best across all tabular datasets due to the heterogeneity in feature space  (See Sections \ref{section_background-preliminaries} and \ref{section_methods-tabular}). Therefore, $N$ methods are ranked from (best) 1 to (worst) N based on their classification performance on each dataset. If two methods have equal weighted F1 scores, the standard deviation breaks the tie. Otherwise, they are assigned the same rank, and the next rank is skipped. No rank is computed for ``OOM'' results. The average rank of each method is obtained across 14 tabular datasets with hard-to-classify data and six datasets with easy-to-classify data.

\textbf {Wilcoxon signed-rank test.} Statistical comparison between two methods (A and B) is not commonly performed in the deep learning literature. We employ the Wilcoxon signed-rank test to gauge statistical significance between the two methods, considering an $\alpha$ value of $0.05$ for statistical significance. This test operates under the assumption of identical sample distributions and avoids the necessity of normality assumptions.

\textbf{Win Matrix.}  We use a win matrix to effectively summarize and better visualize the statistical results from a large pool of methods and datasets. Each cell presents a win ratio in a win matrix determined by tallying dataset instances where method A outperforms method B with statistically distinct outcomes. The result is expressed as a fraction; for instance, `1/8' indicates that A statistically surpasses B on one of eight datasets, whereas B is statistically superior to A on 7/8 dataset instances.

\textbf{Relative F1 score difference.} When comparing F1 scores, we include a metric to determine the percentage of improvement across different methods. We use the percentage of difference in F1 scores as:  
    $ \text{PerDiff}:  \frac{M - m}{M} *100,$
where $M$ and $m$ are the maximum and minimum F1 scores across all the methods for a given dataset.

\section {Results}
\label{section_results}
This section reports experimental results to seek answers to four research questions. First, how do masking and filling strategies in contrastive learning affect the representation learning of tabular data? Second, are contrastive learning methods complementary or superior to attention-based methods for tabular data? Third, when do advanced deep learning methods yield statistically significant improvement over traditional deep and machine learning methods? Fourth, how do data structure and statistics impact a deep model's performance and selection?  All simulations are carried out on a Ubuntu 22.04 machine with Intel(R) Xeon(R) W-2255 CPU @ 3.70GHz CPU with 64GB RAM and Quadro RTX 5000 16GB GPU. The processor has 20 logical cores. PyTorch automatically uses multiple CPUs with parallel processing.  The findings are discussed below in line with the research questions. 

\begin{figure*}[t]
\centering
 \includegraphics[trim=0cm 0cm 0cm 0cm, width=.95\textwidth] {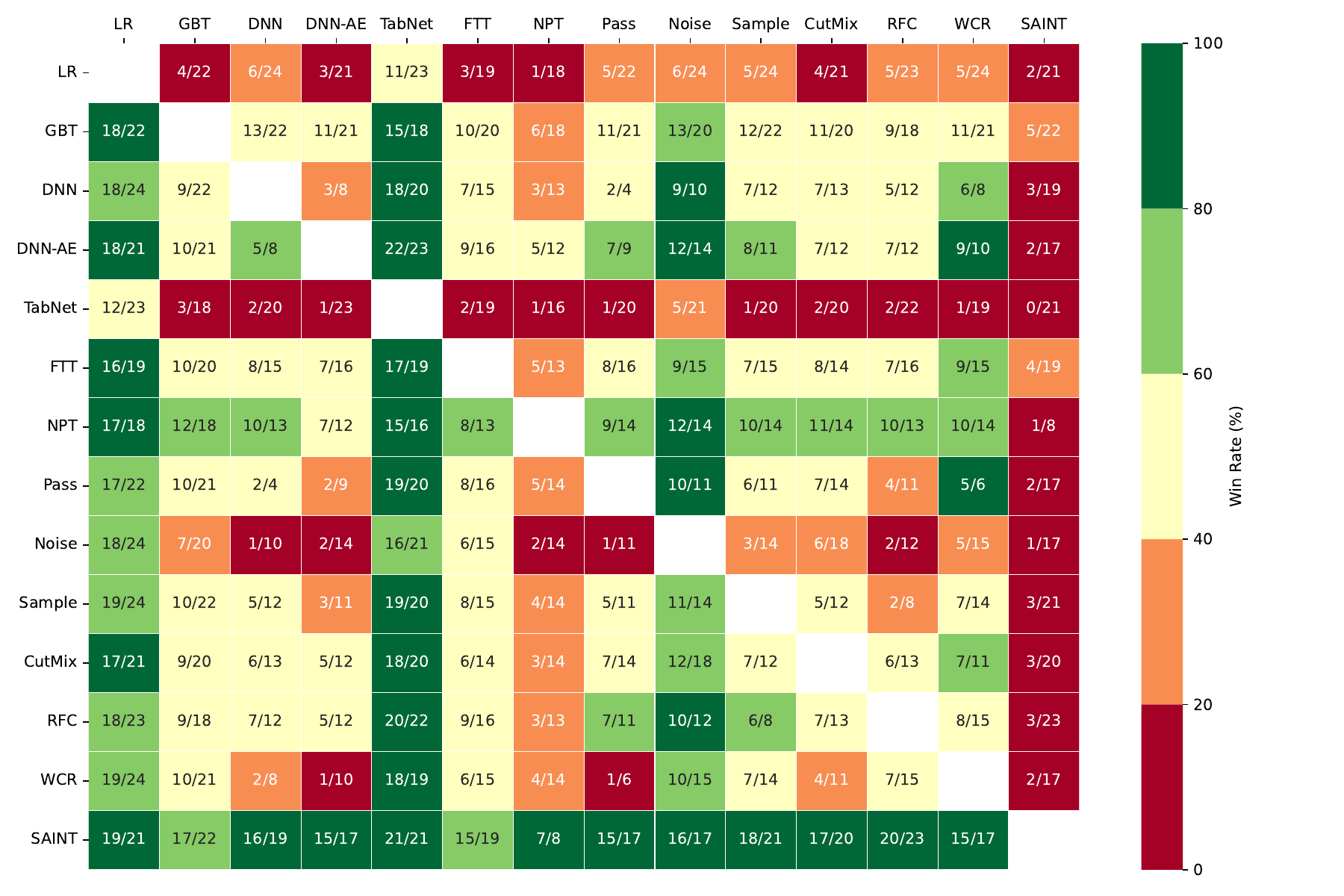}
 \vspace{-10pt}
\caption {Win ratio is presented as the row method against the column method. An x/y ratio indicates that the row method is statistically superior to the column method on x datasets out of y datasets that are statistically significant for the row-column method pair.}
\label{figure_win-matrix}
\end{figure*}

\begin{figure*}[t]
\centering
\vspace{-0pt}
 \includegraphics[trim=0cm 0cm 0cm 0cm, width=.9\textwidth] {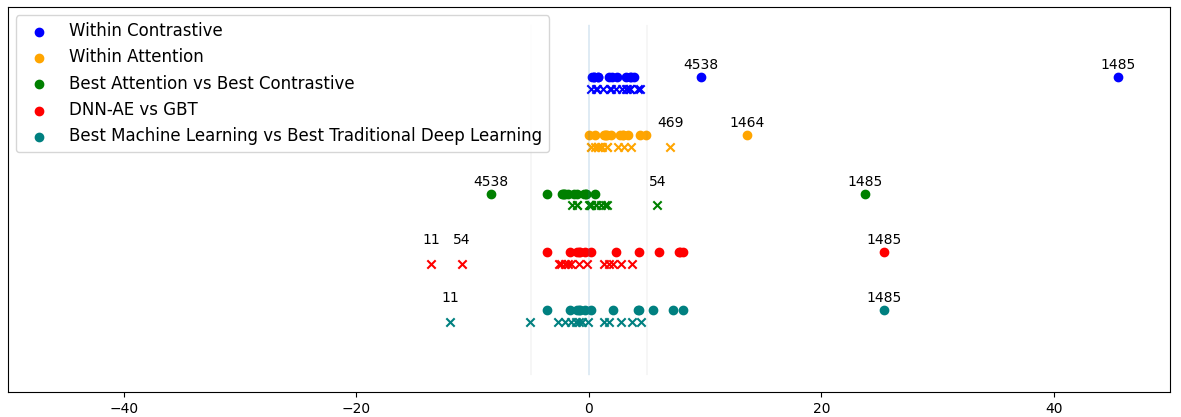}
 \vspace{-5pt}
\caption {Percentage F1 score difference between methods. 
Negative percentages indicate method A outperforms method B in an A versus B comparison. For within-comparison cases, the difference is between the best and worst methods. Hence, the difference is always positive. Full circle markers represent hard datasets, and cross markers represent easy datasets.   }
\label{figu: Relative}
\end{figure*}

\newcolumntype{Z}{>{\arraybackslash}m{2.5em}}
\newcolumntype{Y}{>{\centering\arraybackslash}m{2.5em}}

\begin{table*}
\caption{Rank ordering of machine/deep learning methods across 14 hard-to-classify tabular data sets.}
\vspace{-5pt}
\scalebox{.70}{
\begin{tabular}{ZYY|YYYYYYYYYYYYYY}
\toprule
{OpenML ID} & {F-S Ratio (\%)} & {C-score} & {LR} & {GBT} & {DNN} & {DNN-AE} & {TabNet} & {FTT} & {NPT} & {Pass} & {Noise} & {Sample} & {CutMix} & {RFC} & {WCR} & {SAINT} \\
\midrule
4538 & 0.32 & 0.09 &  14 &  4 &  8 &  11 &  13 &  3 &  2 &  9 &  7 &  10 &  6 &  5 &  12 &  1 \\ \midrule
40975 & 0.35 & 0.05 &  13 &  12 &  10 &  4 &  14 &  1 &  3 &  11 &  9 &  5 &  6 &  7 &  7 &  2 \\ \midrule
40701 & 0.40 & 0.03 &  14 &  3 &  5 &  7 &  11 &  1 &  12 &  9 &  13 &  8 &  4 &  6 &  10 &  2 \\ \midrule
1497 & 0.44 & 0.17 &  14 &  1 &  9 &  10 &  4 &  3 &  5 &  11 &  12 &  8 &  7 &  6 &  13 &  2 \\ \midrule
1464 & 0.54 & 0.47 &  14 &  9 &  12 &  3 &  6 &  13 &  2 &  7 &  11 &  8 &  5 &  10 &  4 &  1 \\ \midrule
23 & 0.61 & 0.13 &  2 &  1 &  12 &  4 &  5 &  13 &  7 &  9 &  11 &  10 &  6 &  14 &  8 &  3 \\ \midrule
1475 & 0.83 & 0.24 &  14 &  1 &  8 &  10 &  13 &  7 &  3 &  11 &  9 &  5 &  6 &  4 &  12 &  2 \\ \midrule
1067 & 1.00 & 0.72 &  14 &  11 &  8 &  6 &  13 &  12 &  1 &  5 &  10 &  7 &  4 &  9 &  3 &  2 \\ \midrule
40982 & 1.39 & 0.25 &  14 &  1 &  9 &  6 &  13 &  2 &  4 &  7 &  8 &  5 &  12 &  10 &  11 &  3 \\ \midrule
1068 & 1.89 & 0.63 &  14 &  3 &  6 &  4 &  13 &  10 &  2 &  8 &  11 &  12 &  5 &  9 &  7 &  1 \\ \midrule
1050 & 2.37 & 0.41 &  14 &  11 &  8 &  7 &  13 &  12 &  2 &  5 &  4 &  9 &  10 &  3 &  6 &  1 \\ \midrule
1049 & 2.54 & 0.35 &  14 &  11 &  8 &  4 &  13 &  10 &  2 &  4 &  12 &  9 &  7 &  3 &  6 &  1 \\ \midrule
1487 & 2.84 & 0.35 &  13 &  11 &  7 &  5 &  11 &  10 &  - &  9 &  6 &  3 &  4 &  2 &  8 &  1 \\ \midrule
1485 & 19.23 & 0.02 &  4 &  2 &  10 &  6 &  9 &  3 &  - &  11 &  12 &  7 &  1 &  5 &  8 &  - \\ \midrule
\multicolumn{3}{c|}{Average} & 12.29 (3.97) & 5.79 (4.66) & 8.57 (1.99) & 6.21 (2.55) & 10.79 (3.40) & 7.14 (4.75) & 3.75 (3.08) & 8.29 (2.40) & 9.64 (2.62) & 7.57 (2.44) & 5.93 (2.67) & 6.64 (3.39) & 8.21 (3.04) & 1.69 (0.75) \\
\bottomrule
\end{tabular}
}
\label{table_rank_hard}
\end{table*}

\begin{table*}
\caption{Rank ordering of machine/deep learning methods across 14 easy-to-classify tabular data sets.}
\vspace{-5pt}
\scalebox{.70}{
\begin{tabular}{ZYY|YYYYYYYYYYYYYY}
\toprule
{OpenML ID} & {F-S Ratio (\%)} & {C-score} & {LR} & {GBT} & {DNN} & {DNN-AE} & {TabNet} & {FTT} & {NPT} & {Pass} & {Noise} & {Sample} & {CutMix} & {RFC} & {WCR} & {SAINT} \\
\midrule
469 & 0.50 & 0.09 &  2 &  3 &  12 &  5 &  14 &  8 &  4 &  9 &  11 &  13 &  6 &  10 &  7 &  1 \\ \midrule
11 & 0.64 & 0.00 &  13 &  14 &  2 &  1 &  12 &  7 &  8 &  4 &  6 &  11 &  9 &  10 &  5 &  3 \\ \midrule
50 & 0.94 & 0.09 &  6 &  5 &  12 &  11 &  14 &  3 &  2 &  8 &  9 &  1 &  13 &  7 &  10 &  4 \\ \midrule
37 & 1.04 & 0.17 &  2 &  1 &  4 &  3 &  5 &  14 &  13 &  6 &  9 &  7 &  12 &  8 &  10 &  11 \\ \midrule
1480 & 1.72 & 0.18 &  14 &  13 &  11 &  9 &  8 &  6 &  2 &  5 &  7 &  4 &  12 &  1 &  10 &  3 \\ \midrule
46 & 1.91 & 0.04 &  3 &  1 &  8 &  4 &  11 &  2 &  - &  9 &  13 &  12 &  6 &  10 &  5 &  7 \\ \midrule
31 & 2.00 & 0.06 &  11 &  10 &  4 &  8 &  14 &  12 &  9 &  6 &  5 &  2 &  13 &  7 &  3 &  1 \\ \midrule
54 & 2.13 & 0.41 &  9 &  14 &  4 &  3 &  13 &  12 &  10 &  6 &  8 &  2 &  1 &  5 &  7 &  11 \\ \midrule
40994 & 3.70 & 0.01 &  14 &  12 &  3 &  6 &  13 &  7 &  9 &  2 &  4 &  8 &  11 &  10 &  5 &  1 \\ \midrule
1494 & 3.89 & 0.17 &  14 &  12 &  7 &  2 &  11 &  13 &  10 &  5 &  8 &  3 &  1 &  6 &  4 &  9 \\ \midrule
1063 & 4.02 & 0.78 &  12 &  5 &  6 &  3 &  14 &  10 &  1 &  8 &  13 &  11 &  7 &  9 &  4 &  2 \\ \midrule
1510 & 5.27 & 0.39 &  11 &  13 &  1 &  5 &  14 &  8 &  12 &  3 &  6 &  9 &  4 &  10 &  2 &  7 \\ \midrule
458 & 8.32 & 0.15 &  7 &  13 &  3 &  3 &  12 &  10 &  - &  8 &  6 &  3 &  1 &  1 &  8 &  11 \\ \midrule
4134 & 47.35 & 0.09 &  3 &  1 &  5 &  2 &  - &  - &  - &  4 &  9 &  7 &  10 &  8 &  6 &  - \\ \midrule
\multicolumn{3}{c|}{Average} & 8.64 (4.72) & 8.36 (5.34) & 5.86 (3.66) & 4.64 (2.92) & 11.92 (2.72) & 8.62 (3.69) & 7.27 (4.27) & 5.93 (2.23) & 8.14 (2.77) & 6.64 (4.14) & 7.57 (4.52) & 7.29 (3.12) & 6.14 (2.63) & 5.46 (4.03) \\
\bottomrule
\end{tabular}
}
\label{table_rank_easy}
\end{table*}

\subsection {Comparing contrastive learning methods}
We first compare different masking and filling strategies proposed for contrastive learning of tabular data in Section \ref{section_contrastive}. 
Tables \ref{table_rank_hard} and \ref{table_rank_easy} show that among six contrastive learning strategies (Pass, Noise, Sample, CutMix, RFC, WCR), CutMix yields the highest F1 scores on eight out of 28 data sets. Among these eight data sets, hard and easy-to-classify datasets are evenly split.  RFC also shows the best F1 scores on eight other data sets, where hard and easy datasets are in a six-to-two ratio. The average rank (AvgRank) in Table \ref{table_results} across 28 data sets reveals that CutMix (AvgRank: 6.75) and RFC (AvgRank: 6.96) are superior to traditional data corruption methods, including adding noise (Noise, AvgRank: 8.89) and sampling from distribution (Sample, AvgRank: 7.11). 

Interestingly, no corruption (Pass, AvgRank: 7.11) performance is on par with the Sample method and superior to the additive noise (Noise) based corruption. This implies that tabular data may need more sophisticated data augmentation or corruption methods beyond adding noise or random sampling. The superior rank ordering of CutMix and RFC holds only on hard datasets (See Table \ref{table_rank_hard} for hard datasets). However, the Pass method yields the best rank (EAvgRank: 5.93) on easy datasets, followed by our proposed WCR (EAvgRank: 6.14), displacing RFC (EAvgRank: 7.29) and CutMix (EAvgRank: 7.57) to fourth and fifth positions. Overall, additive noise is the worst corruption strategy regardless of dataset difficulty.

For most datasets, the percentage difference (PerDiff) between minimum and maximum F1 scores for contrastive learning methods is between $0$ and $5\%$ as shown in Figure \ref{figu: Relative} (within contrastive). Two outlier datasets have shown noticeable improvements (id = 4538, PerDiff: $9.62\%$ and id = 1485, PerDiff: $45.57\%$).  Notably, CutMix has the best F1 score of 0.812 on dataset 1458, substantially outperforming the second-best RFC with an F1 score of 0.607.

The Wilcoxon-signed rank test results presented in the Win matrix (Figure \ref{figure_win-matrix}) show statistical differences between the performances of contrastive learning methods. The win matrix reconfirms CutMix and RFC as the best strategies.  Interestingly, no corruption (Pass) statistically outperforms the noise method on 10/11 datasets and even outperforms the best method (CutMix) on 7/14 datasets. This suggests that contrastive learning is not suitable for all tabular datasets. Our proposed method (WCR) statistically outperforms the best CutMix method on 4/11 and the second-best RFC method on 7/15 datasets. 

\subsection{Attention versus contrastive learning methods}

Among attention-based methods (TabNet, FTT, NPT, and SAINT), only SAINT leverages contrastive learning. Therefore, we evaluate SAINT separately. We first review Tables \ref{table_rank_hard} and \ref{table_rank_easy}, paying attention to the columns ranging from TabNet to WCR that include nine attention-only or contrastive-only methods.

Notably, attention-only methods yield the best F1 scores on 12 out of 14 hard datasets, with NPT taking the lead on seven of these datasets. On the contrary, attention-based methods yield the best F1 scores on only three of 14 easy datasets. This indicates that attention methods are more effective on hard datasets than easy ones. The average rank ordering in the aforementioned tables also supports this claim. The NPT method ranks the best among all contrastive and attention-based methods, averaging 3.75 on hard datasets. In contrast, attention-only methods rank poorly on easy datasets (TabNet rank = 11.92, FFT rank = 8.62, NPT rank = 7.27), securing 9th, 8th, and 4th positions among nine attention or contrastive-based methods. Despite being an attention-based method, TabNet yields the lowest F1 score on 19 out of 28 datasets.

A hybrid method leveraging the strengths of attention and contrastive learning, such as the SAINT method, shows promising improvements in classification performance. SAINT yields the best F1 score among all contrastive and attention-based methods on nine out of 14 hard datasets but only four out of 14 easy ones. However, SAINT has the best average ranks regardless of the dataset difficulty level: 1.69 for hard and 5.46 for easy datasets.

\begin{table*}
\caption{Average F1 scores across 28 tabular data sets. OOM means "Out of Memory". The best-ranked method for each data set is bold-faced. The average rank for each method across 28 datasets is presented at the bottom. In the case of a tie, the F1 score with a lower standard deviation is used as a tie-breaker. Otherwise, we assign the same rank. No rank is considered for "OOM". Pass, Noise, Sample, CutMix, and RFC are corruption methods used in contrastive learning, as described in Section \ref{section_contrastive}. RFC is random feature corruption proposed in the SCARF paper. Diff(\%) shows the relative percentage difference of the best and worst method, i.e., $100 \times \frac{max(scores)-min(scores)}{max(scores)}$, where $scores$ is F1 scores of all methods excluding LR and TabNet for a data set.}
\vspace{-5pt}
\scalebox{.7}{
\label{table_results}
\begin{tabular}{ZYYYYYYYYYYYYYYY}
\toprule
{OpenML ID} & {LR} & {GBT} & {DNN} & {DNN-AE} & {TabNet} & {FTT} & {NPT} & {Pass} & {Noise} & {Sample} & {CutMix} & {RFC} & {WCR} & {SAINT} & {Diff(\%)} \\
\midrule
4538 & 0.447 (0.011) & 0.659 (0.009) & 0.631 (0.011) & 0.607 (0.013) & 0.570 (0.046) & 0.695 (0.009) & 0.715 (0.009) & 0.626 (0.013) & 0.634 (0.013) & 0.613 (0.012) & 0.649 (0.014) & 0.655 (0.012) & 0.592 (0.011) & \bfseries 0.716 (0.008) & 17.32 \\ \midrule
40975 & 0.921 (0.013) & 0.981 (0.011) & 0.988 (0.006) & 0.991 (0.006) & 0.751 (0.021) & \bfseries 1.000 (0.001) & 0.995 (0.006) & 0.987 (0.005) & 0.989 (0.008) & 0.990 (0.007) & 0.990 (0.008) & 0.989 (0.007) & 0.989 (0.007) & 0.999 (0.002) & 1.90 \\ \midrule
40701 & 0.859 (0.011) & 0.947 (0.007) & 0.927 (0.008) & 0.925 (0.010) & 0.916 (0.014) & \bfseries 0.953 (0.005) & 0.911 (0.007) & 0.921 (0.006) & 0.901 (0.009) & 0.923 (0.009) & 0.931 (0.008) & 0.926 (0.007) & 0.921 (0.008) & 0.951 (0.005) & 5.46 \\ \midrule
1497 & 0.700 (0.012) & \bfseries 0.997 (0.002) & 0.925 (0.009) & 0.920 (0.010) & 0.960 (0.007) & 0.968 (0.007) & 0.957 (0.007) & 0.918 (0.010) & 0.914 (0.008) & 0.930 (0.008) & 0.944 (0.007) & 0.948 (0.006) & 0.914 (0.010) & 0.972 (0.005) & 8.32 \\ \midrule
469 & 0.199 (0.025) & 0.194 (0.023) & 0.183 (0.026) & 0.190 (0.031) & 0.178 (0.030) & 0.186 (0.026) & 0.191 (0.028) & 0.185 (0.027) & 0.184 (0.031) & 0.181 (0.024) & 0.189 (0.031) & 0.185 (0.028) & 0.187 (0.030) & \bfseries 0.200 (0.033) & 9.50 \\ \midrule
1464 & 0.676 (0.034) & 0.733 (0.027) & 0.710 (0.030) & 0.760 (0.024) & 0.742 (0.029) & 0.698 (0.024) & 0.773 (0.035) & 0.737 (0.039) & 0.718 (0.029) & 0.736 (0.027) & 0.743 (0.029) & 0.731 (0.032) & 0.745 (0.026) & \bfseries 0.808 (0.028) & 13.61 \\ \midrule
23 & 0.507 (0.028) & \bfseries 0.542 (0.030) & 0.484 (0.029) & 0.498 (0.030) & 0.495 (0.032) & 0.482 (0.025) & 0.493 (0.040) & 0.487 (0.024) & 0.484 (0.024) & 0.485 (0.030) & 0.494 (0.031) & 0.482 (0.031) & 0.490 (0.022) & 0.499 (0.031) & 11.07 \\ \midrule
11 & 0.865 (0.023) & 0.849 (0.019) & 0.981 (0.012) & \bfseries 0.982 (0.012) & 0.891 (0.026) & 0.977 (0.017) & 0.975 (0.019) & 0.979 (0.012) & 0.978 (0.013) & 0.972 (0.015) & 0.974 (0.014) & 0.972 (0.011) & 0.978 (0.012) & 0.980 (0.014) & 13.54 \\ \midrule
1475 & 0.455 (0.016) & \bfseries 0.593 (0.014) & 0.560 (0.012) & 0.557 (0.011) & 0.523 (0.031) & 0.561 (0.013) & 0.579 (0.011) & 0.556 (0.014) & 0.559 (0.012) & 0.563 (0.011) & 0.563 (0.015) & 0.566 (0.011) & 0.544 (0.015) & 0.590 (0.014) & 8.26 \\ \midrule
50 & 0.981 (0.007) & 0.984 (0.012) & 0.966 (0.012) & 0.967 (0.017) & 0.816 (0.043) & 0.995 (0.007) & 0.998 (0.004) & 0.977 (0.014) & 0.975 (0.016) & \bfseries 0.999 (0.002) & 0.955 (0.023) & 0.979 (0.018) & 0.968 (0.016) & 0.990 (0.007) & 4.40 \\ \midrule
1067 & 0.761 (0.016) & 0.831 (0.018) & 0.836 (0.011) & 0.837 (0.014) & 0.826 (0.012) & 0.829 (0.014) & \bfseries 0.854 (0.017) & 0.838 (0.012) & 0.832 (0.013) & 0.837 (0.016) & 0.839 (0.013) & 0.836 (0.016) & 0.839 (0.012) & 0.849 (0.015) & 2.93 \\ \midrule
37 & 0.743 (0.042) & \bfseries 0.749 (0.031) & 0.721 (0.035) & 0.721 (0.032) & 0.720 (0.035) & 0.697 (0.039) & 0.700 (0.043) & 0.720 (0.037) & 0.717 (0.039) & 0.717 (0.034) & 0.706 (0.033) & 0.717 (0.037) & 0.708 (0.038) & 0.708 (0.039) & 6.94 \\ \midrule
40982 & 0.683 (0.025) & \bfseries 0.783 (0.024) & 0.744 (0.023) & 0.749 (0.021) & 0.733 (0.027) & 0.765 (0.016) & 0.755 (0.019) & 0.747 (0.021) & 0.747 (0.024) & 0.749 (0.018) & 0.736 (0.019) & 0.743 (0.023) & 0.740 (0.019) & 0.764 (0.023) & 6.00 \\ \midrule
1480 & 0.650 (0.038) & 0.661 (0.028) & 0.667 (0.037) & 0.671 (0.041) & 0.671 (0.034) & 0.676 (0.041) & 0.683 (0.042) & 0.677 (0.045) & 0.672 (0.038) & 0.677 (0.040) & 0.665 (0.033) & \bfseries 0.687 (0.041) & 0.669 (0.036) & 0.680 (0.046) & 3.78 \\ \midrule
1068 & 0.840 (0.018) & 0.923 (0.011) & 0.920 (0.010) & 0.921 (0.012) & 0.909 (0.013) & 0.918 (0.010) & 0.925 (0.014) & 0.919 (0.012) & 0.915 (0.012) & 0.914 (0.011) & 0.921 (0.014) & 0.919 (0.013) & 0.919 (0.011) & \bfseries 0.932 (0.014) & 1.93 \\ \midrule
46 & 0.953 (0.009) & \bfseries 0.961 (0.008) & 0.945 (0.009) & 0.948 (0.009) & 0.936 (0.012) & 0.956 (0.008) & OOM & 0.944 (0.010) & 0.915 (0.018) & 0.931 (0.011) & 0.946 (0.010) & 0.938 (0.011) & 0.946 (0.009) & 0.945 (0.008) & 4.79 \\ \midrule
31 & 0.716 (0.026) & 0.725 (0.033) & 0.733 (0.020) & 0.726 (0.029) & 0.687 (0.030) & 0.714 (0.023) & 0.725 (0.031) & 0.729 (0.036) & 0.731 (0.027) & 0.735 (0.030) & 0.709 (0.026) & 0.728 (0.026) & 0.734 (0.034) & \bfseries 0.741 (0.030) & 4.32 \\ \midrule
54 & 0.784 (0.031) & 0.736 (0.033) & 0.825 (0.026) & 0.826 (0.025) & 0.765 (0.044) & 0.771 (0.029) & 0.783 (0.030) & 0.821 (0.026) & 0.817 (0.026) & 0.831 (0.024) & \bfseries 0.832 (0.026) & 0.824 (0.024) & 0.820 (0.022) & 0.776 (0.027) & 11.54 \\ \midrule
1050 & 0.813 (0.017) & 0.870 (0.014) & 0.872 (0.013) & 0.873 (0.013) & 0.865 (0.016) & 0.869 (0.012) & 0.887 (0.018) & 0.875 (0.013) & 0.876 (0.016) & 0.872 (0.017) & 0.872 (0.018) & 0.876 (0.013) & 0.874 (0.016) & \bfseries 0.893 (0.014) & 2.69 \\ \midrule
1049 & 0.857 (0.017) & 0.895 (0.019) & 0.906 (0.014) & 0.910 (0.014) & 0.892 (0.027) & 0.903 (0.014) & 0.912 (0.014) & 0.910 (0.014) & 0.892 (0.017) & 0.903 (0.013) & 0.908 (0.016) & 0.910 (0.012) & 0.909 (0.013) & \bfseries 0.916 (0.011) & 2.62 \\ \midrule
1487 & 0.882 (0.010) & 0.927 (0.010) & 0.934 (0.011) & 0.935 (0.010) & 0.927 (0.010) & 0.931 (0.009) & OOM & 0.932 (0.009) & 0.934 (0.009) & 0.935 (0.008) & 0.935 (0.009) & 0.936 (0.009) & 0.933 (0.010) & \bfseries 0.949 (0.009) & 2.32 \\ \midrule
40994 & 0.902 (0.034) & 0.917 (0.026) & 0.942 (0.018) & 0.941 (0.019) & 0.911 (0.021) & 0.934 (0.024) & 0.925 (0.021) & 0.944 (0.020) & 0.942 (0.019) & 0.931 (0.023) & 0.922 (0.020) & 0.925 (0.024) & 0.941 (0.017) & \bfseries 0.949 (0.013) & 3.37 \\ \midrule
1494 & 0.842 (0.029) & 0.854 (0.025) & 0.868 (0.022) & 0.872 (0.021) & 0.854 (0.022) & 0.852 (0.021) & 0.858 (0.020) & 0.870 (0.019) & 0.861 (0.022) & 0.871 (0.023) & \bfseries 0.872 (0.019) & 0.870 (0.021) & 0.870 (0.017) & 0.859 (0.019) & 2.29 \\ \midrule
1063 & 0.797 (0.028) & 0.808 (0.034) & 0.807 (0.027) & 0.815 (0.028) & 0.779 (0.030) & 0.801 (0.026) & \bfseries 0.826 (0.030) & 0.803 (0.028) & 0.791 (0.025) & 0.800 (0.030) & 0.806 (0.032) & 0.802 (0.030) & 0.814 (0.023) & 0.822 (0.027) & 4.24 \\ \midrule
1510 & 0.965 (0.021) & 0.945 (0.018) & \bfseries 0.971 (0.014) & 0.968 (0.017) & 0.938 (0.019) & 0.965 (0.016) & 0.963 (0.018) & 0.970 (0.012) & 0.966 (0.019) & 0.965 (0.017) & 0.969 (0.014) & 0.965 (0.018) & 0.971 (0.016) & 0.965 (0.012) & 2.68 \\ \midrule
458 & 0.995 (0.004) & 0.978 (0.012) & 0.996 (0.004) & 0.996 (0.004) & 0.985 (0.009) & 0.995 (0.007) & OOM & 0.995 (0.005) & 0.996 (0.005) & 0.996 (0.004) & \bfseries 0.997 (0.004) & \bfseries 0.997 (0.004) & 0.995 (0.005) & 0.993 (0.006) & 1.91 \\ \midrule
1485 & 0.612 (0.027) & 0.811 (0.028) & 0.567 (0.023) & 0.605 (0.026) & 0.572 (0.031) & 0.619 (0.034) & OOM & 0.558 (0.024) & 0.442 (0.032) & 0.602 (0.030) & \bfseries 0.812 (0.044) & 0.607 (0.033) & 0.577 (0.022) & OOM & 45.57 \\ \midrule
4134 & 0.767 (0.015) & \bfseries 0.791 (0.013) & 0.765 (0.010) & 0.769 (0.013) & OOM & OOM & OOM & 0.766 (0.012) & 0.750 (0.016) & 0.756 (0.014) & 0.739 (0.017) & 0.755 (0.015) & 0.764 (0.009) & OOM & 6.57 \\ \midrule
Average rank & 10.46 (4.66) & 7.07 (5.09) & 7.21 (3.20) & 5.43 (2.81) & 11.33 (3.09) & 7.85 (4.26) & 5.43 (4.03) & 7.11 (2.57) & 8.89 (2.75) & 7.11 (3.37) & 6.75 (3.74) & 6.96 (3.21) & 7.18 (2.98) & 3.58 (3.43) & - \\
\bottomrule
\end{tabular}
}

\end{table*}

Excluding the worst performing method (TabNet) and dataset id=4134 resulting in the out-of-memory problem, the percentage of difference in maximum and minimum F1 scores within attention-based methods (FTT, NPT, SAINT) ranges from $0\%$ to $5\%$ for most datasets (Figure \ref{figu: Relative}). Two datasets with id = 469 (PerDiff: $7.00\%$) and id = 1464 (PerDiff: $13.61\%$) have produced higher differences in F1 scores. We compare the percentage differences in F1 scores between the best-performing attention method (among TabNet, FTT, NPT) and the best-performing contrastive method (among Pass, Noise, Sample, CutMix, RFC, WCR) for each dataset. Contrastive methods outperform attention on 13 datasets (hard-to-easy split 3:10) where the percentage difference in F1 scores is below $1.61\%$ except for datasets with id = 54 (PerDiff = $5.89\%$) and id = 1485 (PerDiff = $23.77\%$). Attention methods yield higher F1 scores than contrastive on 14 datasets (hard-to-easy split 11:3), out of which seven show more than $1.61\%$, including two large differences on ids = 1464 ($3.62\%$) and 4538 ($8.39\%$). These results generally favor attention over contrastive methods on hard datasets, but for easy data sets, contrastive learning is superior. The dataset with id 1485 results in an out-of-memory problem with the best attention model (NPT). Interestingly, this dataset shows the largest percentage difference in F1 scores between the best and worst contrastive learning methods, which is explained in the discussion section.

The Wilcoxon-signed rank test results presented in the win matrix (Figure \ref{figure_win-matrix}) show that SAINT and NPT statistically outperform contrastive learning methods in most cases. NPT wins with 11/14 and 10/13 scores over CutMix and RFC, respectively, whereas SAINT has 17/20 and 20/23 scores.

On the other hand, SAINT outperforms NPT with a 7/8 score in the win matrix. It is important to note that attention-based methods (NPT, SAINT) run out of memory on five and three datasets, respectively. 
The third attention-based method, FTT, has shown promising performance in previous studies on tabular datasets \citep{Rabbani2023, FTT_Gorishniy2021, han2022adbench, chen2023_tabcaps, yan2023_t2g}. FTT is on par with the best contrastive learning method (cutMix) with an 8/14 score. However, our win matrix shows that FTT loses by a large margin against NPT with a 5/13 score and SAINT (attention plus contrastive) with a 4/19 score.

\begin{table}[t]
\caption{Frequency of datasets under different rank ordering and F-S ratio categories. OOM = out of memory. F-S ratio = feature-to-sample ratio.}
\vspace{-5pt}
\scalebox{0.75}{
\centering
\begin{tabular}{@{}lccccc@{}}
\toprule
   &  \multicolumn{2}{c}{F-S ratio $<$ 2 (16 Datasets)}   &  \multicolumn{3}{c}{F-S ratio $\geq$ 2 (12 Datasets)} \\ 
   \midrule 
   &  1st - 3rd  &  9th - 14th &  1st - 3rd  &  9th - 14th & OOM \\
\midrule
        GBT &  9 & 5 & 2 & 9 & 0 \\
        Contrastive &  3 & N/A & 10 & 0 & 0 \\
        NPT &  8 & 4 & 3 & 5 & 3 \\
        SAINT &  13 & 1 & 5 & 3 & 2 \\
        DNN-AE & 5 & 3 & 3 & 2 & -\\
\bottomrule
\end{tabular}
\label{table:F-S ratio}
}
\vspace{-5pt}
\end{table}

\subsection {Traditional deep versus machine learning methods} 

We compare the performance of traditional deep learning (DNN, DNN-AE) and machine learning methods (LR, GBT) by reviewing Tables \ref{table_results}, \ref{table_rank_hard} and \ref{table_rank_easy}. The latter two tables show GBT as the best on eight of 14 hard datasets with a percentage difference in F1 scores up to 8.11\% (Figure \ref{figu: Relative}) except for dataset id 1458 (PerDiff = 25.4\%). DNN-AE is the best on the remaining six hard datasets, producing up to 3.55\% differences in F1 scores. 

For easy datasets, LR shows the best F1 score only once. GBT and DNN are the best methods on four datasets each. DNN-AE claims the best position on six datasets (tying with DNN on id = 458). Overall, nine of 14 easy datasets are won by traditional deep learning over machine learning methods. We compare the percentage difference in F1 scores between the best of traditional machine learning (e.g., LR or GBT) and the best of traditional deep learning (DNN or DNN-AE) on easy datasets. The percentage differences in favor of machine learning are in the range between $1.35\%$ and $4.52\%$, whereas the ones in favor of deep learning are between $0.1\%$ and $2.65\%$ except in datasets with id = 11 (PerDiff = $5.08\%$) and id = 54 (PerDiff = $11.91\%$). 

DNN-AE, DNN, and GBT show similar performances in the win matrix, with DNN having the lowest overall win scores. A comparison between GBT and DNN-AE reveals that the former outperforms the latter on 11/21 datasets, which is a tie. LR ranks the worst among the four traditional methods regardless of the dataset difficulty. However, LR is still effective on several datasets as it scores several points against GBT (4/22), DNN (6/24), and DNN-AE (3/21). The overall best method (SAINT) statistically outperforms all traditional methods by a large margin - GBT (17/22), DNN (15/17), and DNN-AE (15/17).

\begin{table}[t]
\caption{Frequency of datasets under different rank ordering and C-score categories. C-score = Mean absolute correlation of feature pairs.}
\vspace{-0pt}
\scalebox{0.75}{
\centering
\begin{tabular}{@{}lcccc@{}}
\toprule
 & \multicolumn{2}{c}{C-score $\leq$ 0.1 (11 Datasets)} & \multicolumn{2}{c}{C-score $\geq$ 0.3 (9 Datasets)} \\ \midrule
 & 1st - 3rd & 9th - 14th & 1st - 3rd & 9th - 14th \\ \midrule
GBT & 5 & 4 & 1 & 7 \\
Contrastive & 4 & 0 & 6 & 0 \\
NPT & 3 & 3 & 6 & 4 \\
SAINT & 7 & 0 & 7 & 1 \\ 
DNN-AE & 2 & 2 & 3 & 0 \\
\bottomrule
\end{tabular}
\label{table:C-score}}
\vspace{-15pt}
\end{table}

\subsection{Effects of tabular data structure and statistics}

It is evident that the performance of learning algorithms is very specific to data sets. It warrants further investigation into the effects of data structure and statistics on model performance and selection.  Our comparison in this section is based on the rank ordering of the methods because F1 scores largely vary between datasets.

The model performance shows some interesting relationships with the feature-sample (F-S) ratio of the dataset. We refer the reader to Tables \ref{table_rank_hard} and \ref{table_rank_easy} with attention to the F-S column in ascending order. A summary of the results below is presented in Table \ref{table:F-S ratio}. GBT excels on datasets with low F-S ratios (tall tabular data matrix), showcasing top ranks, but experiences a decline in performance with increasing F-S ratios. In particular, GBT ranks between the 9th and 14th positions on the 12 datasets with F-S ratio $\geq$ 2, except for three datasets (ids = 1063, 1485, and 4134). One extreme observation includes datasets 1485 and 4134 with the largest F-S ratios (19.2 and 47). GBT ranks second and first on these datasets, respectively. These two datasets have a very high dimensional and uncorrelated feature space with low C-scores (0.02 and 0.09). To support this conjecture, we observe that GBT ranks the best method on the only other dataset (id = 46) with more than 100 features (287 features after one-hot encoding) and a low C-score (0.04). For 16 datasets with F-S ratios $<$ 2, GBT ranks within the 9th to 14th positions in five cases (ids = 40975, 1464, 11, 1067, 1480), but between the 1st and 3rd positions for nine datasets.

Conversely, contrastive methods achieve better rank orderings on datasets with high F-S ratios (high data dimensionality) than those with low F-S ratios. For datasets with F-S ratios $\geq$ 2, there is always at least one contrastive strategy that ranks between the 1st and 3rd positions, with the exceptions of datasets with IDs 1063 and 4134 (best rank 4). On the contrary, for F-S ratios $<$ 2, contrastive methods rank among the top three for only three out of 16 cases (ids = 50, 37, 1480).

NPT struggles mostly on high F-S ratio datasets, while SAINT shows a decrease in performance. Among datasets with an F-S ratio $\geq$ 2, SAINT ranks worse than nine on three (NPT - five) out of twelve datasets. SAINT encounters OOM problems on two (NPT - three) occasions. Nevertheless, SAINT ranks the best on five datasets. For datasets with F-S ratios $<$ 2, SAINT ranks within the top three on 13 (NPT - eight)  of 16 datasets. For this F-S ratio range, SAINT is worse than rank nine on only one dataset (id = 37), which is true for NPT in only four out of 16 cases.

Table \ref{table:C-score} shows the rank frequency of different methods for high and low C-score dataset groups. Contrastive and SAINT appear to perform strongly regardless of C-scores. However, GBT performs relatively worse when the C-score is high (more correlated features). DNN-AE is observed to be relatively better on data sets with high C-scores but is evenly split when C-scores are low.

SAINT is one of the best two methods on every hard dataset, except for IDs 12 and 40982 (rank 3) and 1485 (OOM - no rank), with the best average rank of 1.69. NPT ranks between the first and third positions on 8 out of 14 hard datasets and has the second-best average rank of 3.75. This solidifies the superiority of the attention-based approach on hard datasets. Attention-based methods yield worse average ranks (SAINT rank: 5.46, NPT rank: 7.27) on easy data sets.

\subsection {Time complexity}

Time complexity is an essential factor in benchmarking learning algorithms. Table \ref{table_time} presents the total run time for training and testing on 30 bootstrapped samples of the dataset with ID 1510. Notably, GBT is the most computationally efficient method, requiring orders of magnitude less time than any other approach. For example, the run time of DNN is 30 times higher than that of GBT, whereas FTT is almost 100 times slower than GBT. Traditional deep learning methods (DNN and DNN-AE) are among the second-fastest (894 and 1120 seconds), while contrastive and attention-based methods take the third and fourth spots. FTT offers a good time complexity and performance trade-off, being as competitive as the best contrastive learning method but two times faster. 


\begin{table}[]
\caption{Total time taken in seconds to train and test each model on  30 bootstrapped samples from the dataset with ID 1510.}
\vspace{-5pt}
\centering
\scalebox{0.9}{
\begin{tabular}{lrr}
\toprule
  Method &  Time/seconds &  log(time) \\
\midrule
        LR &               6 &                 0.796 \\
        GBT &              37 &                 1.569 \\
        DNN &             894 &                 2.951 \\
     DNN-AE &            1120 &                 3.049 \\
        FTT &            3407 &                 3.532 \\
      Pass &            6951 &                 3.842 \\
      Noise &            7059 &                 3.848 \\
      CutMix &            7040 &                 3.847 \\
     Sample &            7165 &                 3.855 \\
         RF &            7118 &                 3.852 \\
     TabNet &           10394 &                 4.016 \\
     SAINT &            15129&                 4.180 \\
        NPT &           27978 &                 4.446 \\
\bottomrule
\end{tabular}
\label{table_time}
}
\vspace{-10pt}
\end{table}

\section {Discussion}
\label{section_discussion}
This paper investigates the effectiveness of recent deep learning breakthroughs (attention and contrastive learning) in tabular data representations when deep learning has not achieved much success against traditional machine learning. The findings of the paper are summarized as follows.  First, tabular data need more sophisticated data augmentation or corruption methods than sampling or 
additive noise to yield the benefit of contrastive learning. The contrastive learning approach, which is most effective on image data, is not preferred for tabular data. While cutMix is the most effective contrastive learning method for tabular data,  contrastive learning is not recommended for 
easy-to-classify datasets. Second, attention-based methods are superior to contrastive learning on tabular datasets. Leveraging the strength of attention and contrastive learning yields the best performance by far (SAINT), followed by NPT and FTT methods. The best contrastive method performs on par with the third-best attention-based method (FTT). However, attention-based methods fail, resulting in out-of-memory issues on datasets with large data dimensionality, including datasets 1485 and 4134 with F-S ratio 19.2\% and 47.3\% respectively, which is in line with a previous study~\citep{Rabbani2023}. Third, recent breakthroughs in deep learning (attention plus contrastive) have shown a strong and consistent performance boost to conquer "the castle" of traditional machine learning that has been dominating the tabular data domain. Fourth, contrastive learning appears superior to other methods when the number of features is more than twice the number of samples. Fifth, the best-performing deep learning methods (attention-based) should not be considered for tabular data with a large number of features ($>$100) because these methods are vulnerable to high data dimensionality and out-of-memory issues. Contrastive learning methods should be chosen instead for high-dimensional tabular data sets.

\subsection{Data set-specific model selection}

The wide variation of model performance across tabular data sets is not unknown in practice. Contrary to this notion, the AI literature often presents a new method that is often found to outperform all baselines on all datasets. While this may be true for image or text data, similar monolithic observation is rare in the tabular data literature. Benchmark datasets are often selectively chosen (e.g., the ones with large sample size, low dimensionality, or simulated easy-to-classify datasets) in studies to demonstrate the superiority of a proposed method, disregarding the effect or conditions of data structure and statistics on the model performance.

The current practice in data science tests a data set on a variety of classifiers to identify the best model. In contrast to this arbitrary trial, we argue that the structure and statistics of the data set should guide the selection of the classifier. For example, many benchmarking datasets with an easy-to-classify decision boundary should not be tested on a sophisticated and computationally expensive deep model when a traditional machine-learning method yields top-notch accuracy within a few seconds. When a sophisticated model is prone to overfit on such easy-to-classify data, hurting the overall rank ordering of a promising method. In reality, tabular datasets similar to many popular samples used for benchmarking purposes (e.g., Iris, breast cancer, wine quality) are few and far between. A classic example in computer vision is the widespread use of the twenty-five-year-old MNIST dataset, on which countless methods have achieved over 99\% accuracy. Therefore, stretching the MNIST binary digit classification accuracy by another 0.01\%, introducing a hefty deep model is not in the greater interest of machine learning research. Having some prior knowledge of the dataset is imperative before selecting and justifying a model for classification or representation learning. Future methods should focus on a targeted group of datasets under challenging and meaningful criteria instead of proposing one generalized model to show its best performance regardless of datasets and data domains.

\subsection{Easy-versus-hard dataset considerations}

Taking advantage of low computational costs, traditional machine learning can be used to promptly reveal the complexity of a dataset. A contrast between the classification accuracies of logistic regression and non-linear gradient boosting tree (GBT) classifiers should hint if the dataset is worth exploring on advanced deep methods. An insignificant contrast between LR and GBT accuracies hints at a relatively simpler decision boundary. Our analysis on easy-to-classify datasets reveals in Table~\ref{table_rank_easy} that traditional deep learning with an autoencoder-based pretraining (DNN-AE) yields the best average rank (4.64 (2.92)), which is up to 20 times faster and also superior to attention and contrastive learning based advanced deep methods. 

On the other hand, when GBT performs strongly against LR on hard datasets, it suggests the presence of a complex decision boundary. This scenario replicates the common observation in the literature that traditional machine learning (average rank ordering of GBT 5.79 (4.66)) outperforms deep learning methods (Rank orderings of DNN (8.57 (1.99) and DNN-AE (6.21 (2.55)), as shown in Table~\ref{table_rank_hard}. In response to this notion, an attention-based model (NPT: average rank 3.75 (3.08)) and an attention-contrastive hybrid model (SAINT: average rank 1.69 (0.75) present consistently superior performance, challenging the current dominance of traditional machine learning on tabular data. The SAINT method shows a good trade-off in performance on easy and hard datasets by incorporating the benefits of contrastive and attention-based learning but requiring less computational time than the attention-only method (NPT). 

\subsection{Contrastive learning is still useful}

Our win matrix shows that no contrastive method beats GBT, and it even ties with computationally six times faster DNN-AE. Despite the mediocre performance,  contrastive learning is sometimes preferred over other superior methods. The best-performing methods (NPT, SAINT) are vulnerable to out-of-memory problems when feature dimensionality is close to 100 and above. For high-dimensional tabular datasets, contrastive learning methods are preferred for hard datasets with complex decision boundaries.

One interesting observation is made on data set 1485, which stands out as an outlier in most experimental scenarios in Figure~\ref{figu: Relative}. This is a hard dataset with high feature dimensionality (500) and an almost null C-score (0.02).  A previous study has shown that this dataset has low feature correlations, which resulted in a GBT classification accuracy substantially higher ($>$15\%) than those reported for LR or deep learning methods (DNN or DNN-AE)~\citep{Perturbation_Abrar2022}. Contrastive learning (cutMix) is the only method that outperforms GBT (F1 scores 81.2 versus 81.1). In contrast, attention-based methods (NPT, SAINT) completely fail (out-of-memory) and other methods barely achieve a 60\% F1 score in Table~\ref{table_results}. In several easy datasets with low C-scores (ID 11 (C-score = 0) and 40994 (C-score = 0.01)), attention-based methods (FTT and NPT) struggle, while contrastive methods like Pass perform better. The hybrid method SAINT excels even further. The success of SAINT in these situations could be attributed to its ability to switch to contrastive learning when an attention-based approach falls short.

\subsection{Future research directions}
We identify four research directions to improve the current state of tabular data learning. First, it is imperative to investigate explainable metrics to quantify and qualify a tabular data set for specific models. The statistical underpinning of why certain datasets are more suitable for a specific model may help build data-specific models instead of one that fits all. For example, several data sets, including 458, fail to take the advantage of pretraining because DNN produces the same performance as DNN-AE. Therefore, which data sets may not reap the benefit of pre-training needs further investigation. Second, 
despite superior performances in many cases, high computational costs and poor explainability of deep learning models remain two major roadblocks toward their practical applications. Future methods are expected to improve computational costs and explainability of deep tabular data learning. Third, attention-based models need some major improvement in handling high-dimensional tabular data without demanding a large sample size. Fourth, innovative solutions to data corruption or augmentation methods for tabular data may advance contrastive learning of tabular data.

\section{Conclusions}\label{section_conclusions}
The paper presents the first extensive benchmarking of 
attention and contrastive learning methods on tabular data. The findings of this article substantiate our claim that the rank ordering of a learning algorithm widely varies due to the heterogeneity of tabular datasets, which necessitates more data-specific interpretation and innovation of learning algorithms. Therefore, choosing data sets and baselines can introduce selection bias in benchmarking tabular data learning methods unlike standards followed in the computer vision literature. With sufficient computing resources and some limits on data dimensionality, attention-based methods are significantly superior to traditional machine learning methods. There is still room to improve the computational efficiency of attention-based methods and innovative methods for contrastive learning of tabular data. Complementing the strengths of attention and contrastive learning can pave the path for more accurate and computationally efficient methods for tabular data in the future. 



\section*{Acknowledgment}

The research reported in this publication was supported by the Air Force Office of Scientific Research under Grant Number W911NF-23-1-0170. The content is solely the responsibility of the authors and should not be interpreted as representing the official policies, either expressed or implied, of
the Army Research Office or the U.S. Government.


\bibliographystyle{unsrt}
\bibliography{mybib}

\begin{thebibliography}{10}

\bibitem{Maksims2023}
Maksims Kazijevs and Manar~D. Samad.
\newblock Deep imputation of missing values in time series health data: A
  review with benchmarking.
\newblock {\em Journal of Biomedical Informatics}, 144:104440, 2023.

\bibitem{Grinsztajn2022}
Leo Grinsztajn, Edouard Oyallon, and Gael Varoquaux.
\newblock Why do tree-based models still outperform deep learning on typical
  tabular data?
\newblock In {\em Thirty-sixth Conference on Neural Information Processing
  Systems Datasets and Benchmarks Track}, 2022.

\bibitem{Borisov2022}
Vadim Borisov, Tobias Leemann, Kathrin Se{\ss}ler, Johannes Haug, Martin
  Pawelczyk, and Gjergji Kasneci.
\newblock Deep neural networks and tabular data: A survey.
\newblock {\em IEEE Transactions on Neural Networks and Learning Systems},
  2022.

\bibitem{Kadra2021}
Arlind Kadra, Marius Lindauer, Frank Hutter, and Josif Grabocka.
\newblock Well-tuned simple nets excel on tabular datasets.
\newblock {\em Advances in neural information processing systems},
  34:23928--23941, 2021.

\bibitem{Dua_2019}
Dheeru Dua and Casey Graff.
\newblock {UCI} machine learning repository, 2017.

\bibitem{ICACI_Abrar2023}
Sakib Abrar, Ali Sekmen, and Manar~D. Samad.
\newblock Effectiveness of deep image embedding clustering methods on tabular
  data.
\newblock In {\em 2023 15th International Conference on Advanced Computational
  Intelligence (ICACI)}, pages 1--7, 2023.

\bibitem{Wang2021}
Zhenhua Wang, Olanrewaju Akande, Jason Poulos, and Fan Li.
\newblock {Are deep learning models superior for missing data imputation in
  large surveys? Evidence from an empirical comparison}.
\newblock 2021.

\bibitem{Hamori2018}
Shigeyuki Hamori, Minami Kawai, Takahiro Kume, Yuji Murakami, and Chikara
  Watanabe.
\newblock Ensemble learning or deep learning? application to default risk
  analysis.
\newblock {\em Journal of Risk and Financial Management}, 11(1):12, 2018.

\bibitem{Kohler2019}
Niklas~D K{\"o}hler, Maren B{\"u}ttner, and Fabian~J Theis.
\newblock Deep learning does not outperform classical machine learning for
  cell-type annotation.
\newblock {\em BioRxiv}, page 653907, 2019.

\bibitem{Smith2020}
Aaron~M Smith, Jonathan~R Walsh, John Long, Craig~B Davis, Peter Henstock,
  Martin~R Hodge, Mateusz Maciejewski, Xinmeng~Jasmine Mu, Stephen Ra, Shanrong
  Zhao, et~al.
\newblock Standard machine learning approaches outperform deep representation
  learning on phenotype prediction from transcriptomics data.
\newblock {\em BMC bioinformatics}, 21(1):1--18, 2020.

\bibitem{Shwartz-Ziv2022}
Ravid Shwartz-Ziv and Amitai Armon.
\newblock Tabular data: Deep learning is not all you need.
\newblock {\em Information Fusion}, 81:84--90, 5 2022.

\bibitem{Huang2020Fusion}
Shih~Cheng Huang, Anuj Pareek, Saeed Seyyedi, Imon Banerjee, and Matthew~P.
  Lungren.
\newblock {Fusion of medical imaging and electronic health records using deep
  learning: a systematic review and implementation guidelines}.
\newblock {\em npj Digital Medicine 2020 3:1}, 3(1):1--9, oct 2020.

\bibitem{Gavito2023}
Andrea~Trevi{\~{n}}o Gavito, Diego Klabjan, and Jean Utke.
\newblock {Multi-Layer Attention-Based Explainability via Transformers for
  Tabular Data}.
\newblock feb 2023.

\bibitem{Kim2023}
Minwook Kim, Juseong Kim, Jose Bento, and Giltae Song.
\newblock {Revisiting Self-Training with Regularized Pseudo-Labeling for
  Tabular Data}.
\newblock {\em arXiv preprint arXiv:2302.14013}, feb 2023.

\bibitem{dataPreProcess01}
Steve Lohr.
\newblock For big-data scientists,‘janitor work’is key hurdle to insights.
\newblock {\em New York Times}, 17:B4, 2014.

\bibitem{Hancock2020}
John~T. Hancock and Taghi~M. Khoshgoftaar.
\newblock {Survey on categorical data for neural networks}.
\newblock {\em Journal of Big Data}, 7(1):28, dec 2020.

\bibitem{FTT_Gorishniy2021}
Yury Gorishniy, Ivan Rubachev, Valentin Khrulkov, and Artem Babenko.
\newblock Revisiting deep learning models for tabular data.
\newblock {\em Advances in Neural Information Processing Systems},
  23:18932--18943, 2021.

\bibitem{NPT_Kossen2021}
Jannik Kossen, Neil Band, Clare Lyle, Aidan~N Gomez, Thomas Rainforth, and
  Yarin Gal.
\newblock Self-attention between datapoints: Going beyond individual
  input-output pairs in deep learning.
\newblock {\em Advances in Neural Information Processing Systems},
  34:28742--28756, 2021.

\bibitem{Vaswani2017}
Ashish Vaswani, Noam Shazeer, Niki Parmar, Jakob Uszkoreit, Llion Jones,
  Aidan~N Gomez, {\L}ukasz Kaiser, and Illia Polosukhin.
\newblock Attention is all you need.
\newblock {\em Advances in neural information processing systems}, 30, 2017.

\bibitem{ResNet_he2016}
Kaiming He, Xiangyu Zhang, Shaoqing Ren, and Jian Sun.
\newblock Deep residual learning for image recognition.
\newblock In {\em Proceedings of the IEEE conference on computer vision and
  pattern recognition}, pages 770--778, 2016.

\bibitem{AutoInt_Song2019}
Weiping Song, Chence Shi, Zhiping Xiao, Zhijian Duan, Yewen Xu, Ming Zhang, and
  Jian Tang.
\newblock {AutoInt: Automatic Feature Interaction Learning via Self-Attentive
  Neural Networks}.
\newblock In {\em Proceedings of the 28th ACM International Conference on
  Information and Knowledge Management}, pages 1161--1170, New York, NY, USA,
  nov 2019. ACM.

\bibitem{TabTransformer_Huang2020}
Xin Huang, Ashish Khetan, Milan Cvitkovic, and Zohar Karnin.
\newblock {TabTransformer: Tabular Data Modeling Using Contextual Embeddings}.
\newblock {\em arXiv preprint arXiv:2012.06678}, dec 2020.

\bibitem{GrowNet_badirli2022}
Sarkhan Badirli, Xuanqing Liu, Zhengming Xing, Avradeep Bhowmik, Khoa~D Doan,
  and Sathiya Keerthi.
\newblock Gradient boosting neural networks: Grownet, 2022.

\bibitem{NODE_Popov2020}
Sergei Popov, Stanislav Morozov, and Artem Babenko.
\newblock Neural oblivious decision ensembles for deep learning on tabular
  data.
\newblock In {\em International Conference on Learning Representations}, 2020.

\bibitem{Perturbation_Abrar2022}
Sakib Abrar and Manar~D Samad.
\newblock Perturbation of deep autoencoder weights for model compression and
  classification of tabular data.
\newblock {\em Neural Networks}, 156:160--169, 2022.

\bibitem{SimCLR_Chen_2020}
Ting Chen, Simon Kornblith, Mohammad Norouzi, and Geoffrey Hinton.
\newblock A simple framework for contrastive learning of visual
  representations.
\newblock In {\em International conference on machine learning}, pages
  1597--1607. PMLR, 2020.

\bibitem{SCARF_Bahri2021}
Dara Bahri, Heinrich Jiang, Yi~Tay, and Donald Metzler.
\newblock Scarf: Self-supervised contrastive learning using random feature
  corruption.
\newblock In {\em International Conference on Learning Representations}, 2022.

\bibitem{OpenMLCC18}
Bernd Bischl, Giuseppe Casalicchio, Matthias Feurer, Pieter Gijsbers, Frank
  Hutter, Michel Lang, Rafael~Gomes Mantovani, Jan~N. van Rijn, and Joaquin
  Vanschoren.
\newblock Open{ML} benchmarking suites.
\newblock In {\em Thirty-fifth Conference on Neural Information Processing
  Systems Datasets and Benchmarks Track (Round 2)}, 2021.

\bibitem{cutmix_yun2019}
Sangdoo Yun, Dongyoon Han, Seong~Joon Oh, Sanghyuk Chun, Junsuk Choe, and
  Youngjoon Yoo.
\newblock Cutmix: Regularization strategy to train strong classifiers with
  localizable features.
\newblock In {\em Proceedings of the IEEE/CVF international conference on
  computer vision}, pages 6023--6032, 2019.

\bibitem{mixup_zhang2017}
Hongyi Zhang, Moustapha Cisse, Yann~N. Dauphin, and David Lopez-Paz.
\newblock mixup: Beyond empirical risk minimization.
\newblock In {\em International Conference on Learning Representations}, 2018.

\bibitem{Darabi2021}
Sajad Darabi, Shayan Fazeli, Ali Pazokitoroudi, Sriram Sankararaman, and Majid
  Sarrafzadeh.
\newblock {Contrastive Mixup: Self- and Semi-Supervised learning for Tabular
  Domain}.
\newblock {\em arXiv preprint arXiv:2108.12296}, aug 2021.

\bibitem{SAINT_Somepalli2022}
Gowthami Somepalli, Avi Schwarzschild, Micah Goldblum, C~Bayan Bruss, and Tom
  Goldstein.
\newblock Saint: Improved neural networks for tabular data via row attention
  and contrastive pre-training.
\newblock In {\em NeurIPS 2022 First Table Representation Workshop}, 2022.

\bibitem{Iscen_2019_label_propagation}
Ahmet Iscen, Giorgos Tolias, Yannis Avrithis, and Ondrej Chum.
\newblock Label propagation for deep semi-supervised learning.
\newblock In {\em Proceedings of the IEEE/CVF Conference on Computer Vision and
  Pattern Recognition (CVPR)}, June 2019.

\bibitem{STab_Hajiramezanali2022}
Ehsan Hajiramezanali, Nathaniel Diamant, Gabriele Scalia, and Max~W {Shen
  Genentech}.
\newblock {STab: Self-supervised Learning for Tabular Data}.
\newblock In {\em NeurIPS 2022 First Table Representation Workshop}, 2022.

\bibitem{SubTab_Ucar2021}
Talip Ucar, Ehsan Hajiramezanali, and Lindsay Edwards.
\newblock Subtab: Subsetting features of tabular data for self-supervised
  representation learning.
\newblock {\em Advances in Neural Information Processing Systems},
  34:18853--18865, 2021.

\bibitem{Chen2023_ExcelFormer}
Jintai Chen, Jiahuan Yan, Danny~Ziyi Chen, and Jian Wu.
\newblock {ExcelFormer: A Neural Network Surpassing GBDTs on Tabular Data}.
\newblock jan 2023.

\bibitem{hollmann2022_tabpfn}
Noah Hollmann, Samuel M{\"u}ller, Katharina Eggensperger, and Frank Hutter.
\newblock Tab{PFN}: A transformer that solves small tabular classification
  problems in a second.
\newblock In {\em The Eleventh International Conference on Learning
  Representations}, 2023.

\bibitem{chen2023_tabcaps}
Jintai Chen, Kuanlun Liao, Yanwen Fang, Danny~Z Chen, and Jian Wu.
\newblock {TabCaps: A Capsule Neural Network for Tabular Data Classification
  with BoW Routing}.
\newblock In {\em The Eleventh International Conference on Learning
  Representations}, sep 2023.

\bibitem{yan2023_t2g}
Jiahuan Yan, Jintai Chen, Yixuan Wu, Danny~Z Chen, and Jian Wu.
\newblock T2g-former: organizing tabular features into relation graphs promotes
  heterogeneous feature interaction.
\newblock In {\em Proceedings of the AAAI Conference on Artificial
  Intelligence}, volume~37, pages 10720--10728, 2023.

\bibitem{TabNet_Arik2021}
Sercan~O. Arik and Tomas Pfister.
\newblock Tabnet: Attentive interpretable tabular learning.
\newblock pages 6679--6687, 8 2021.

\bibitem{Bahdanau2014}
Dzmitry Bahdanau, Kyunghyun Cho, and Yoshua Bengio.
\newblock {Neural Machine Translation by Jointly Learning to Align and
  Translate}.
\newblock In {\em ICLR}, sep 2014.

\bibitem{cho2014rnn}
Kyunghyun Cho, Bart van Merri{\"e}nboer, Caglar Gulcehre, Dzmitry Bahdanau,
  Fethi Bougares, Holger Schwenk, and Yoshua Bengio.
\newblock Learning phrase representations using {RNN} encoder{--}decoder for
  statistical machine translation.
\newblock In Alessandro Moschitti, Bo~Pang, and Walter Daelemans, editors, {\em
  Proceedings of the 2014 Conference on Empirical Methods in Natural Language
  Processing ({EMNLP})}, pages 1724--1734, Doha, Qatar, October 2014.
  Association for Computational Linguistics.

\bibitem{luong2015globalAttention}
Thang Luong, Hieu Pham, and Christopher~D. Manning.
\newblock Effective approaches to attention-based neural machine translation.
\newblock In Llu{\'\i}s M{\`a}rquez, Chris Callison-Burch, and Jian Su,
  editors, {\em Proceedings of the 2015 Conference on Empirical Methods in
  Natural Language Processing}, pages 1412--1421, Lisbon, Portugal, September
  2015. Association for Computational Linguistics.

\bibitem{Chaudhari2021attentionSurvey1}
Sneha Chaudhari, Varun Mithal, Gungor Polatkan, and Rohan Ramanath.
\newblock {An Attentive Survey of Attention Models}.
\newblock {\em ACM Transactions on Intelligent Systems and Technology}, 12(5),
  oct 2021.

\bibitem{Niu2021attentionSurvey2}
Zhaoyang Niu, Guoqiang Zhong, and Hui Yu.
\newblock {A review on the attention mechanism of deep learning}.
\newblock {\em Neurocomputing}, 452:48--62, sep 2021.

\bibitem{Dosovitskiy2020Vit}
Alexey Dosovitskiy, Lucas Beyer, Alexander Kolesnikov, Dirk Weissenborn,
  Xiaohua Zhai, Thomas Unterthiner, Mostafa Dehghani, Matthias Minderer, Georg
  Heigold, Sylvain Gelly, Jakob Uszkoreit, and Neil Houlsby.
\newblock An image is worth 16x16 words: Transformers for image recognition at
  scale.
\newblock {\em CoRR}, abs/2010.11929, 2020.

\bibitem{devlin2018bert}
Jacob Devlin, Ming-Wei Chang, Kenton Lee, and Kristina Toutanova.
\newblock Bert: Pre-training of deep bidirectional transformers for language
  understanding.
\newblock In {\em North American Chapter of the Association for Computational
  Linguistics}, 2019.

\bibitem{grill2020bootstrap}
Jean-Bastien Grill, Florian Strub, Florent Altch{\'e}, Corentin Tallec, Pierre
  Richemond, Elena Buchatskaya, Carl Doersch, Bernardo Avila~Pires, Zhaohan
  Guo, Mohammad Gheshlaghi~Azar, et~al.
\newblock Bootstrap your own latent-a new approach to self-supervised learning.
\newblock {\em Advances in neural information processing systems},
  33:21271--21284, 2020.

\bibitem{chen2020improved}
Xinlei Chen, Haoqi Fan, Ross Girshick, and Kaiming He.
\newblock Improved baselines with momentum contrastive learning.
\newblock {\em arXiv preprint arXiv:2003.04297}, 2020.

\bibitem{tian2020goodviews}
Yonglong Tian, Chen Sun, Ben Poole, Dilip Krishnan, Cordelia Schmid, and
  Phillip Isola.
\newblock What makes for good views for contrastive learning?
\newblock {\em Advances in neural information processing systems},
  33:6827--6839, 2020.

\bibitem{chen2020big}
Ting Chen, Simon Kornblith, Kevin Swersky, Mohammad Norouzi, and Geoffrey~E
  Hinton.
\newblock Big self-supervised models are strong semi-supervised learners.
\newblock {\em Advances in neural information processing systems},
  33:22243--22255, 2020.

\bibitem{zbontar2021barlow}
Jure Zbontar, Li~Jing, Ishan Misra, Yann LeCun, and St{\'e}phane Deny.
\newblock Barlow twins: Self-supervised learning via redundancy reduction.
\newblock In {\em International Conference on Machine Learning}, pages
  12310--12320. PMLR, 2021.

\bibitem{he2020momentum}
Kaiming He, Haoqi Fan, Yuxin Wu, Saining Xie, and Ross Girshick.
\newblock Momentum contrast for unsupervised visual representation learning.
\newblock In {\em Proceedings of the IEEE/CVF conference on computer vision and
  pattern recognition}, pages 9729--9738, 2020.

\bibitem{wang2020understanding}
Tongzhou Wang and Phillip Isola.
\newblock Understanding contrastive representation learning through alignment
  and uniformity on the hypersphere.
\newblock In {\em International Conference on Machine Learning}, pages
  9929--9939. PMLR, 2020.

\bibitem{VIME_Yoon2020}
Jinsung Yoon, Yao Zhang, James Jordon, and Mihaela {Van Der Schaar}.
\newblock {VIME: Extending the Success of Self-and Semi-supervised Learning to
  Tabular Domain}.
\newblock In {\em Advances in Neural Information Processing Systems 33}, 2020.

\bibitem{Rubachev2022}
Ivan Rubachev, Artem Alekberov, Yury Gorishniy, and Artem Babenko.
\newblock Revisiting pretraining objectives for tabular deep learning.
\newblock {\em arXiv preprint arXiv:2207.03208}, 2022.

\bibitem{MIDA_Gondara2018}
Lovedeep Gondara and Ke~Wang.
\newblock {MIDA: Multiple imputation using denoising autoencoders}.
\newblock In {\em Lecture Notes in Computer Science (including subseries
  Lecture Notes in Artificial Intelligence and Lecture Notes in
  Bioinformatics)}, volume 10939 LNAI, pages 260--272. Springer Verlag, 2018.

\bibitem{TABBIE_Iida2021}
Hiroshi Iida, Dung Thai, Varun Manjunatha, and Mohit Iyyer.
\newblock {TABBIE: Pretrained Representations of Tabular Data}.
\newblock In {\em Proceedings of the 2021 Conference of the North American
  Chapter of the Association for Computational Linguistics: Human Language
  Technologies}, pages 3446--3456, Stroudsburg, PA, USA, 2021. Association for
  Computational Linguistics.

\bibitem{Scheff2016}
Stephen~W. Scheff.
\newblock {\em {Fundamental Statistical Principles for the Neurobiologist: A
  Survival Guide}}.
\newblock Elsevier, jan 2016.

\bibitem{Samad2019JAAC}
Manar~D Samad, Alvaro Ulloa, Gregory~J Wehner, Linyuan Jing, Dustin Hartzel,
  Christopher~W Good, Brent~A Williams, Christopher~M Haggerty, and Brandon~K
  Fornwalt.
\newblock {Predicting Survival From Large Echocardiography and Electronic
  Health Record Datasets: Optimization With Machine Learning}.
\newblock {\em JACC: Cardiovascular Imaging}, 12(4):681--689, jun 2019.

\bibitem{Rabbani2023}
Shourav~B Rabbani and Manar~D Samad.
\newblock Between-sample relationship in learning tabular data using graph and
  attention networks.
\newblock {\em ICDATA Las Vegas}, 2023.

\bibitem{han2022adbench}
Songqiao Han, Xiyang Hu, Hailiang Huang, Minqi Jiang, and Yue Zhao.
\newblock Adbench: Anomaly detection benchmark.
\newblock {\em Advances in Neural Information Processing Systems},
  35:32142--32159, 2022.

\end{thebibliography}

\end{document}